\documentclass[preprint,5p,twocolumn,times,authoryear]{elsarticle}

\usepackage{amssymb}
\usepackage{amsmath,amsfonts}

\usepackage{lineno}

\graphicspath{ {./Figures/} }

\usepackage{caption}
\usepackage{subcaption}

\usepackage[table]{xcolor}
\newcommand*{\new}[1]{#1}

\usepackage{hyperref}

\usepackage{algorithm}
\usepackage{algpseudocode}

\journal{International Journal of Applied Earth
Observation and Geoinformation}

\begin{document}

\begin{frontmatter}

\title{Individual Tree Detection in Large-Scale Urban Environments using High-Resolution Multispectral Imagery}

\author[cs]{Jonathan Ventura}
\author[bio]{Camille Pawlak}
\author[soc]{Milo Honsberger}
\author[soc]{Cameron Gonsalves}
\author[cs]{Julian Rice}
\author[bio]{Natalie L.R. Love}
\author[cs]{Skyler Han}
\author[cs]{Viet Nguyen}
\author[soc,poli]{Keilana Sugano}
\author[bus]{Jacqueline Doremus}
\author[soc]{G. Andrew Fricker}
\author[bio]{Jenn Yost}
\author[bio]{Matt Ritter}

\affiliation[cs]{organization={Department of Computer Science \& Software Engineering, California Polytechnic State University},addressline={1 Grand Ave},city={San Luis Obispo},postcode={93107-0354},state={CA},country={USA}}
\affiliation[bio]{organization={Department of Biological Sciences, California Polytechnic State University},addressline={1 Grand Ave},city={San Luis Obispo},postcode={93107-0401},state={CA},country={USA}}
\affiliation[soc]{organization={Department of Social Sciences, California Polytechnic State University},addressline={1 Grand Ave},city={San Luis Obispo},postcode={93107-0329},state={CA},country={USA}}
\affiliation[poli]{organization={Department of Political Science, California Polytechnic State University},addressline={1 Grand Ave},city={San Luis Obispo},postcode={93107-0329},state={CA},country={USA}}
\affiliation[bus]{organization={Orfalea College of Business, California Polytechnic State University},addressline={1 Grand Ave},city={San Luis Obispo},postcode={93107},state={CA},country={USA}}

\begin{abstract}
Systematic maps of urban forests are useful for regional planners and ecologists to understand the spatial distribution of trees in cities.  However, manually-created urban forest inventories are expensive and time-consuming to create and typically don't provide coverage of private land.  Toward the goal of automating urban forest inventory through machine learning techniques, we performed a comparative study of methods for automatically detecting and localizing trees in  multispectral aerial imagery of urban environments, and introduce a novel method based on convolutional neural network regression.  Our evaluation is supported by a new dataset of over 1,500 images and almost 100,000 tree annotations, covering eight cities, six climate zones, and three image capture years.  \new{Our method outperforms previous methods, achieving 73.6\% precision and 73.3\% recall when trained and tested in Southern California, and 76.5\% precision 72.0\% recall when trained and tested across the entire state.  To demonstrate the scalability of the technique, we produced the first map of trees across the entire urban forest of California. The map we produced provides important data for the planning and management of California's urban forest, and establishes a proven methodology for potentially producing similar maps nationally and globally in the future.}
\end{abstract}


\end{frontmatter}

\section{Introduction}

Urban forests provide extensive benefits to residents of cities such as contributing to resident energy-savings, reducing impervious runoff, improving water quality, controlling microclimate, and sequestering carbon \citep{mcpherson2002comparison,mcpherson2016structure,livesley2016urban}. City, state, or county governments manage the public section in order to estimate and maximize those benefits and make sure they are equitably distributed throughout a city.  Governments typically track their urban forests through manual tree inventories performed by professional arborists \citep{nielsen2014review}.  However, such inventories are typically limited to information about public street trees, which make up only a portion of a city's urban forest.  The extent of the benefits provided to residents by their urban trees depends on the number of trees in that city, both public and private.  When making policy decisions about where the next tree planting is needed most, cities need to be able to account for both the publicly and privately managed urban forest in order to target plantings. Automatic tree detection implemented using machine learning and aerial imagery provides a method for cities to efficiently and cost-effectively track their urban forests.

Tree detection methods can be divided into two categories: methods that operate on LiDAR-derived products such as a canopy height model (CHM) \citep{chen20092d,silva_imputation_2016,liu_novel_2015,wu_individual_2016,zorner2018lidar,roussel2020lidr,xu2021crown,munzinger2022mapping} or a 3D point cloud \citep{ayrey_use_2018,chen_individual_2021}, and methods that operate solely on optical imagery.  The geometric information obtained from LiDAR is highly useful for tree detection.  However, LiDAR is expensive to collect, and there is currently no source for high-resolution LiDAR data for much of the Earth, unlike high-resolution multi-spectral imagery, for which multiple sources exist.

\new{The advantages of using optical, multispectral imagery are that it is collected everywhere in the United States as part of the U.S. National Agricultural Imagery Program (NAIP) and it is cheaper to collect than LiDAR.  LiDAR has the benefit of precisely measuring vertical dimensions, but coverage is spotty with unreliable repeat times.  The primary motivation for using NAIP imagery is its wall-to-wall coverage and repeated collections which would allow for tracking changes over time.  Indeed, a blended approach using optical and LiDAR data would be preferable, but is not repeatable at scale.}

Early methods for tree detection in remote sensing imagery perform per-pixel classification to identify tree canopies, tree characteristics, and tree species \citep{xiao_using_2004,yang2009tree,jensen_classification_2012,alonzo_identifying_2013,shang_classification_2014,bosch2020detectree,brandt2020unexpectedly}.  The limitation of most of these approaches is that they output per-pixel maps rather than detecting and localizing individual trees.  However, some methods have been proposed to extract individual tree locations from the per-pixel maps as a post-processing step, such as template matching \citep{yang2009tree} and connected components \citep{brandt2020unexpectedly}.

Many recent methods on tree detection adopt an object detection approach \citep{santos2019assessment,zamboni2021benchmarking,weinstein2019individual,zhang2022individual,das2022geoai,beloiu2023individual} or an instance segmentation approach \citep{martins2021deep,freudenberg2022individual,yang2022detecting,sun2022counting,ball2023accurate}.  Object detection requires a bounding box annotation for each tree in the training dataset, while instance segmentation requires an accurate delineation of the tree crown.  Bounding box and crown delineation annotations have size information which provides a useful supervisory signal for machine learning methods.  However, fully annotating a bounding box or complete crown delineation for each tree is time-consuming, and it is often difficult for human annotators to visually determine the correct delineation of overlapping tree crowns.

In this study we focus on the alternative approach of annotating and predicting tree location points.  This approach makes the annotation process easier and more scalable.  However, object detection and instance segmentation methods are unable to learn from point annotations alone because they lack size information.  Instead, we can consider methods from the substantial literature on object counting \citep{lempitsky2010learning}, where the goal is to learn to count the number of objects in an image from only point annotations.   Some previous work \citep{osco2020convolutional,chen2022transformer} has been successful in adapting these techniques for tree detection.  In this study we expand upon the method of \cite{osco2020convolutional} and introduce several modifications to improve the results.

\new{Since our method only outputs point locations for the trees, our results cannot be used to directly estimate tree coverage area or tree crown size.  However, the tree count estimates produced by our method would be generally useful to city managers, urban foresters, and ecologists.  For example, when developing management plans, we believe city managers value tree counts over canopy estimates for estimating maintenance hours and staff requirements \citep{dwyer1992assessing}.  Furthermore, ecologists could use our tree count estimates to track patterns in the urban forest over space and time \citep{love2022diversity}.}

\new{The primary objectives of our research are to produce a methodology for tree detection in urban areas in California and create a map of tree locations for every urban tree in California. In this work, we explore the potential for automatic detection and localization of trees in the California urban forest from aerial imagery. Using machine learning methods to detect trees in imagery, we can automatically produce a map of the trees in an urban environment, covering both public and private land and providing geographic coordinates for each tree.}

To the best of our knowledge, no previous study has evaluated the potential for machine learning-based methods to automate urban tree detection across the entire state of California.  Previous studies have either evaluated urban tree detection in individual cities in California \citep{wegner2016cataloging,branson2018google}
or have addressed related but different tasks, such as tree detection in the natural forests of California \citep{weinstein2019individual,chen2022transformer} or automated tree species identification from aerial imagery of California cities \citep{beery2022auto}.  Our dataset covers several cities across California, a state with considerable topographic and climatic diversity, and also includes imagery from multiple years to evaluate the potential for longitudinal analysis.  Specifically, our hand-annotated dataset spans eight cities, six climate zones, and three image capture years, and including almost 100,000 tree annotations in total.  The high value, broad impact, and diversity of California’s urban forest make it such that studies done in this region are broadly applicable to urban forests globally.





\section{Study Area and Dataset}

\subsection{Study Area}

California is the third largest state in the U.S. and the most populous with a population of 	39,538,223 according to the 2020 Decennial Census \citep{USCensusBureau2020}. The California urban forest provides services and benefits to California's residents previously estimated to be worth \$8.3 billion annually \citep{mcpherson2017structure}. California's urban forests are composed of species with wide-ranging climatic requirements \citep{love2022diversity}. California's cities can host these species because the state has substantial topographic and climatic diversity, with six distinct California climate zones covering deserts, coastal and alpine environments \citep{love2022diversity, clarkediversity}.  The California urban forest has high diversity compared to other urban forests, with a higher diversity ranking than the United States at a national scale, and a higher ranking relative to the most diverse urban forests within the United States, the Western region \citep{love2022diversity,madiversity}. California’s urban forests are so diverse that they are comparable not just to urban forests, but to the most diverse forests in the world, tropical forests \citep{love2022diversity}. 

\subsection{Dataset}

We prepared our training and testing datasets using imagery from U.S. National Agricultural Program (NAIP), which contains aerial imagery acquired during the agricultural growing seasons in the United States.  NAIP multispectral imagery is acquired every two years and covers the entire contiguous United States, typically at 1 m resolution or 60 cm resolution.  We chose NAIP imagery because its coverage, temporal and spatial resolution, and public availability make it an ideal resource to support large-scale study of the urban forest in the United States.   In our study we only used 60 cm imagery collected in 2016 or later.

\begin{figure}
    \centering
    \includegraphics[width=0.9\columnwidth]{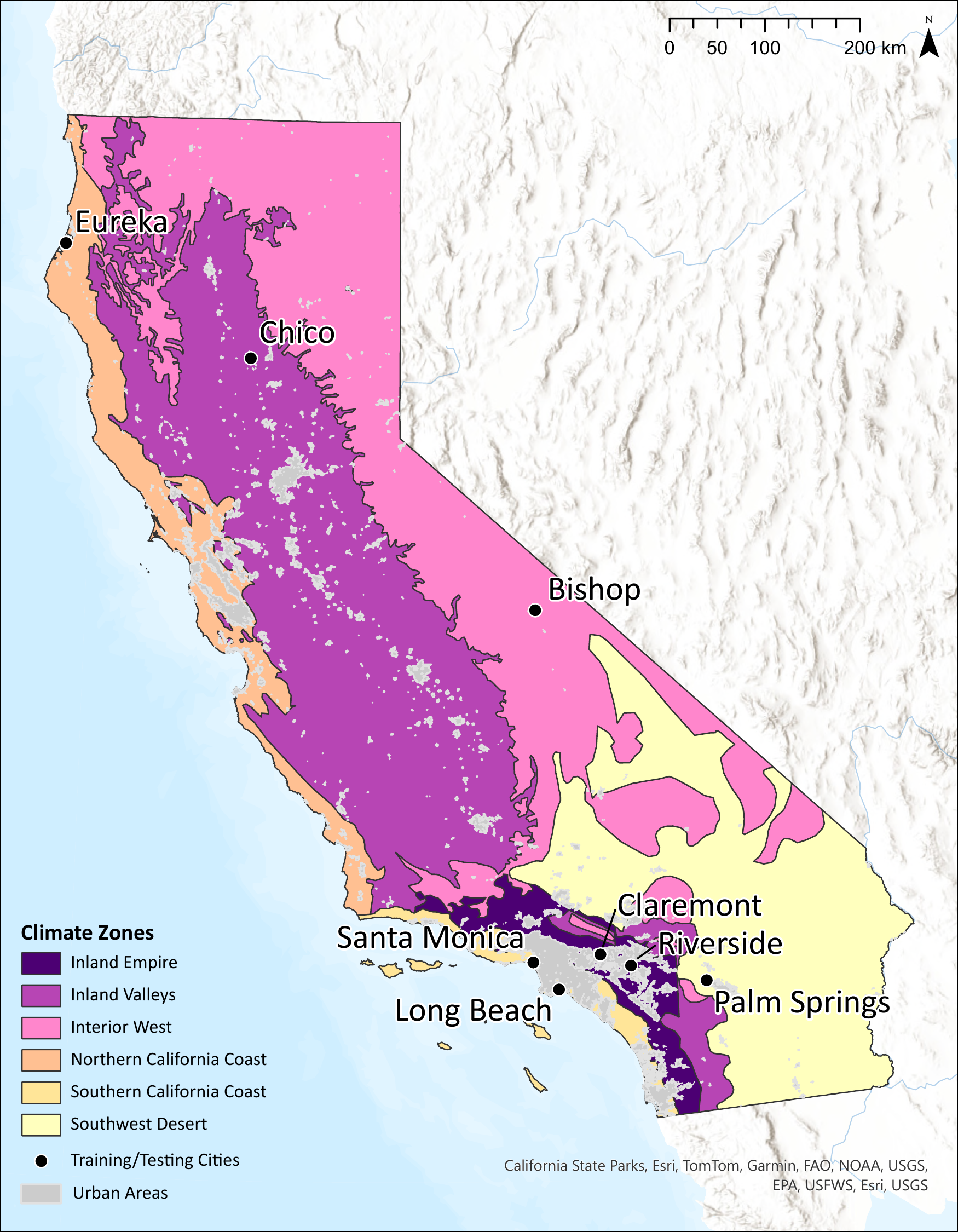}
    \caption{Locations of cities from which we collected and annotated images to form our dataset.  Each climate zone in California is represented by at least one city in the dataset. \new{The climate zones represent the following proportions of the total urban reserve area across the state: Inland Empire (24.03\%), Inland Valleys (26.36\%). Interior West (2.25\%), Northern California Coast (16.46\%), Southern California Coast (24.65\%), and Southwest Desert (6.25\%).}}
    \label{fig:california_map}
\end{figure}

Table \ref{tab:dataset_summary} summarizes our dataset and Figure \ref{fig:california_map} illustrates the locations of the included cities and climate zones.  In total, our dataset contains 1,651 images and 96,425 annotated trees, and covers eight cities and all six climate zones in California \cite{mcpherson2010selecting}.

The Southern California 2020 portion of the dataset covers five cities in Southern California and contains roughly 90-100 square image tiles captured in 2020 from each city: Claremont, Long Beach, Palm Springs, Riverside, and Santa Monica.  Each image has a size of $256\times256$ pixels, a resolution of 60 cm, and includes red, green, blue, and near infrared channels.  We collected and annotated NAIP imagery from 2020 for these sites.  We withheld a random 10\% split of the Southern California 2020 images for testing and used the remaining 90\% for training.

To further test the extrapolation ability of the network, we prepared two additional subsets of imagery.  We annotated the same sites in Southern California in the Southern California 2020 portion of the dataset, but using imagery from 2016 and 2018, to test the ability of our method to process imagery of a similar location but captured at different times.  We also selected and annotated images from three cities in Northern California: Bishop, Chico, and Eureka.  These cities are in different climate zones from the Southern California sites, and in some places have a higher density of trees, since the cities include more densely forested rural areas.  Thus, these images serve to test the ability of our method to extrapolate to other regions different from the training sites.

\new{Our complete dataset combines the Southern and Northern California subsets and data from all three years (2016, 2018, and 2020).  The dataset covers all five of California's climate zones, with an emphasis on the Southern California Coast and Inland Empire, which contain 48.68\% of the state's total urban area.  After developing our initial model using the Southern California 2020 subset and validating its extrapolation ability with the remaining images, we re-trained the model using the complete dataset to obtain our final tree detection model.}

\new{The various dataset subsets used in the paper are summarized in Table \ref{tab:subsets}, which lists the number of images and tree annotations in the train and test splits for each subset.}

We used tree inventories acquired by the cities as a starting point to annotate by hand the locations of all trees in the images, including trees in both public and private spaces.  Since the city inventories only included trees on public land, we added points where necessary to ensure coverage of all trees visible in the imagery, including trees on private land such as in front yards, backyards, and parking lots.  We also visually checked the accuracy of each point and moved it if necessary to ensure that it was on top of the tree trunk location in the image.  If we couldn't determine the tree trunk location by visual inspection, we put the point on the center of the canopy.  This happened sometimes with palm trees, for example, which have thin trunks that are easily confused with their shadow.  We compared with imagery at a higher spatial resolution where available to verify tree locations and to ensure that no non-trees (such as shrubs) were included in the annotations.

\begin{table}[t]
    \centering
    \caption{Summary of our dataset of annotated NAIP tiles in California.  The number of trees listed in the three right-most columns is the number of hand-annotated trees in the study area in each city and in each year. A dash indicates that the area was not annotated for that year.  In total, our dataset contains 95,972 tree annotations.  The climate zone abbreviations are Inland Empire (IE), Inland Valleys (IV), Interior West (IW), Northern California Coast (NCC), Southern California Coast (SCC),
 and Southwest Desert (SD).  The cities above the line are in Southern California, and below are in Northern California.  The cells highlighted in gray indicate the Southern California 2020 portion of the dataset.}
    \resizebox{\columnwidth}{!}{
    \begin{tabular}{llrrrr}
          &  &  & \multicolumn{3}{c}{\textbf{Tree Annotations}} \\
         \textbf{City} & \textbf{Zone} & \textbf{Images} &  \textbf{2016} & \textbf{2018} & \textbf{2020}  \\
         \hline
         Claremont & IE & 92 & 4,856 & 4,794 & \cellcolor[HTML]{D3D3D3}4,668  \\
         Long Beach & SCC & 100 & 6,470 & 6,402 & \cellcolor[HTML]{D3D3D3}5,843 \\
         Palm Springs & SD  & 100 & 4,431 & 4,704 & \cellcolor[HTML]{D3D3D3}4,107 \\
         Riverside  & IE & 90 & 5,015 & 4,399 & \cellcolor[HTML]{D3D3D3}4,082 \\
         Santa Monica & SCC & 92 & 5,822 & 5,829 & \cellcolor[HTML]{D3D3D3}5,841 \\
         \hline
         Bishop  & IW & 10 & - & - & 682  \\
         Chico  & IV & 99 & - & 8,185 & 8,162  \\
         Eureka  & NCC & 21 & - & - & 2,133  \\
         \hline
         Total & & 1,651 & 26,594 & 34,313 & 35,518
    \end{tabular}
}
    \label{tab:dataset_summary}
\end{table}

\begin{table}[t]
    \centering
    \caption{\new{Summary of dataset subsets used in our experiments.}}
    \new{
    \resizebox{\columnwidth}{!}{
    \begin{tabular}{lrrrr}
    & \multicolumn{2}{c}{\textbf{Images}} & \multicolumn{2}{c}{\textbf{Annotations}} \\
    \textbf{Subset} & \textbf{Train} & \textbf{Test} &  \textbf{Train} & \textbf{Test} \\
         \hline
         Southern California 2020 & 426 & 48 & 21,861 & 2,680 \\
         Southern California 2016-2018 & -  & 948 & - & 52,722 \\
         Northern California 2018-2020 & - & 229 & - & 19,162 \\
         Complete dataset & 1,485 & 166 & 87,666 & 8,759 
    \end{tabular}
}
}
    \label{tab:subsets}
\end{table}

\section{Methods}
\label{sec:methods}

\begin{figure*}[t]
    \centering
     \begin{subfigure}[b]{0.32\textwidth}
         \centering
    \includegraphics[width=\textwidth]{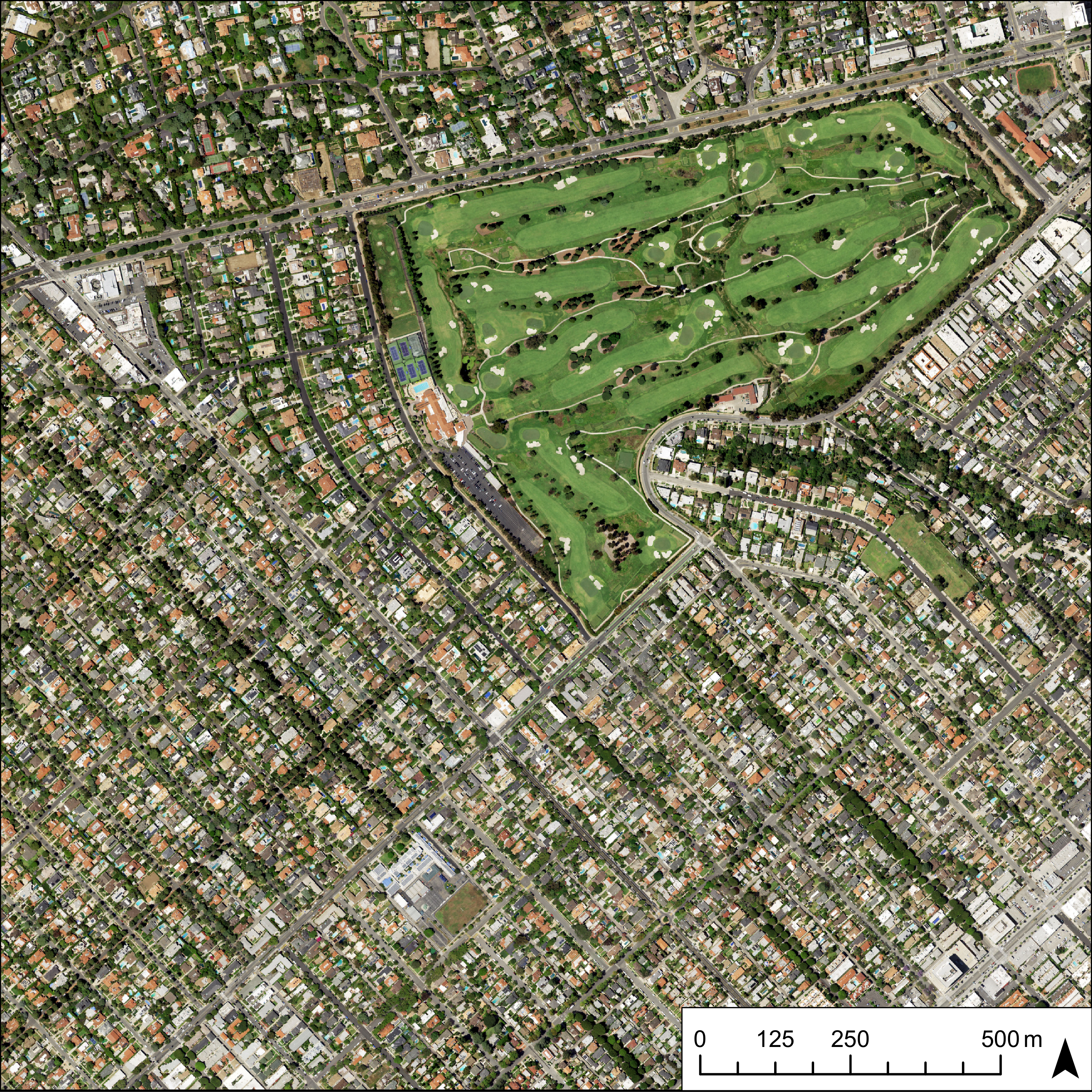}
        \caption{Input raster}
        \label{subfig:input_raster}
     \end{subfigure}
     \hfill
     \begin{subfigure}[b]{0.32\textwidth}
        \centering
        \includegraphics[width=\textwidth]{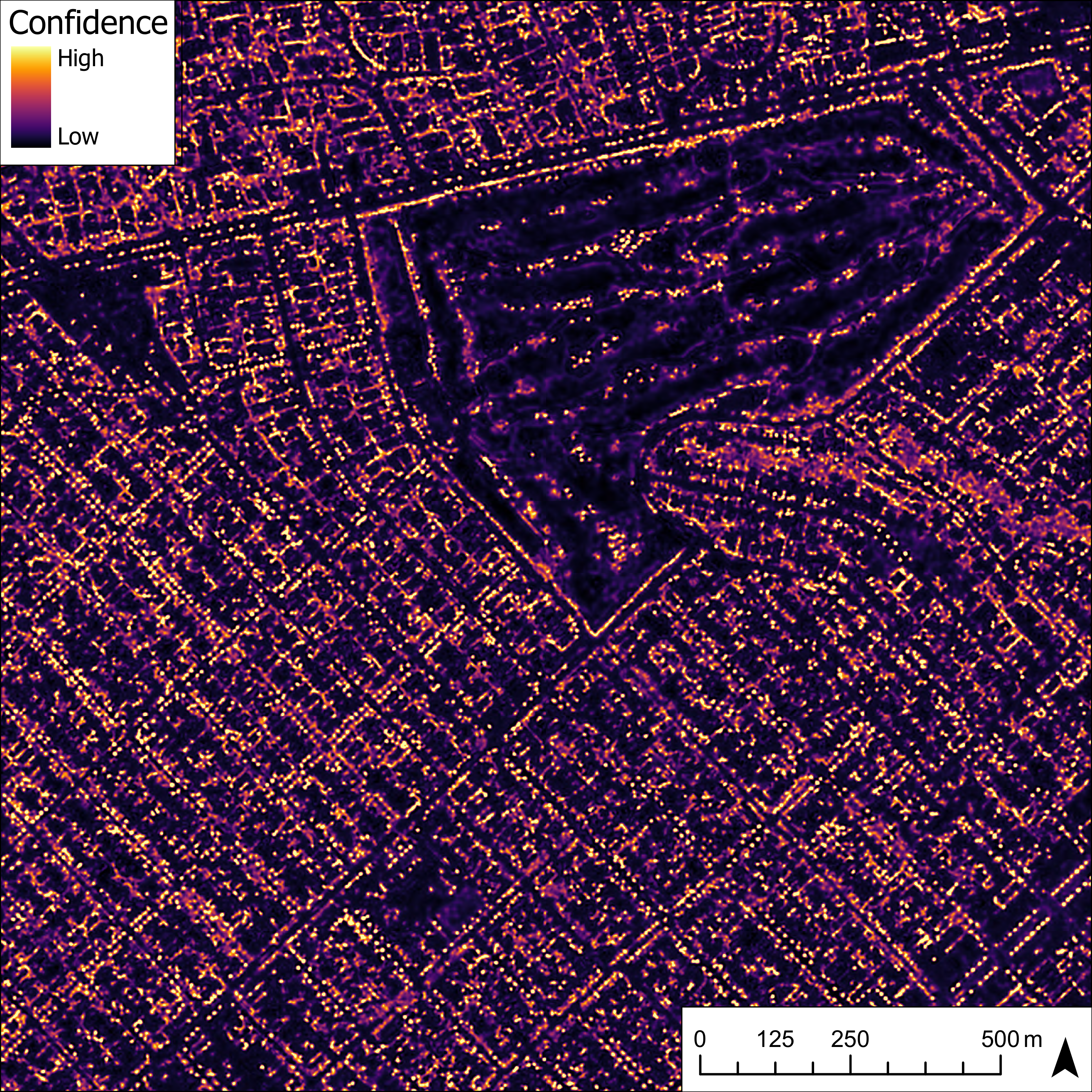}
        \caption{Output confidence map}
        \label{subfig:density_map}
    \end{subfigure}
    \hfill
    \begin{subfigure}[b]{0.32\textwidth}
        \centering
        \includegraphics[width=\textwidth]{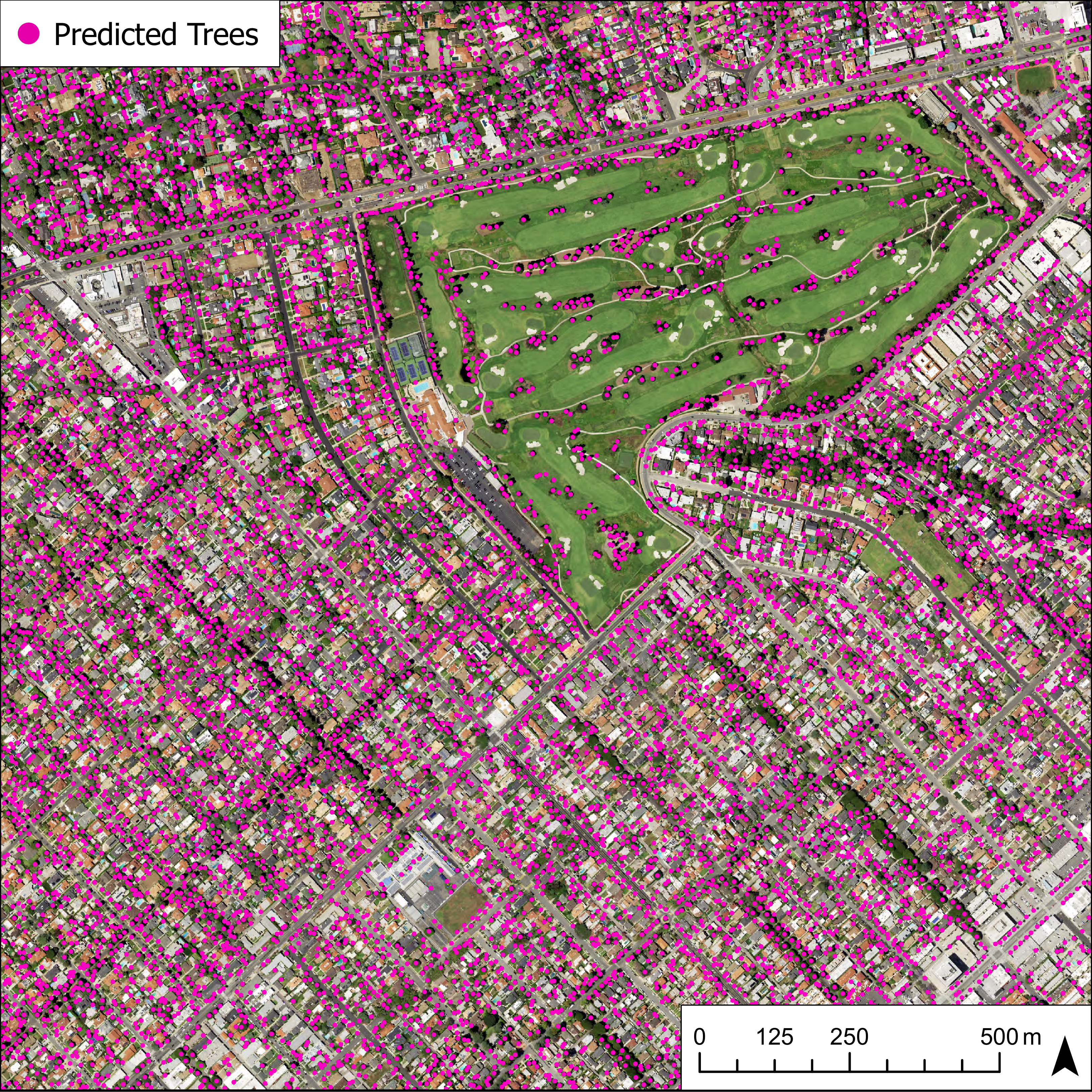}
        \caption{Detected trees}
        \label{subfig:detected_trees}
    \end{subfigure}
    \caption{Example of tree detection using our method on a section of \new{2020} NAIP imagery in Santa Monica, CA.  We process the input raster (a) in a CNN to produce a confidence map (b).  We then apply peak finding in the confidence map to produce individual tree detections (c).   }
    \label{fig:inference}
\end{figure*}


We introduce a novel tree detection method using neural network confidence map regression.  Because our annotations are tree location points, we found a confidence map approach \citep{lempitsky2010learning,osco2020convolutional} to be the most appropriate for our task.  Our overall process for tree detection and localization is illustrated in Figure \ref{fig:inference}.  The input to the system is a stack of raw and derived rasters.  The raster stack is passed through a CNN which outputs a single-channel confidence map.  Peaks in the confidence map should correspond to tree locations.  We identify tree locations in the confidence map using local peak finding (non-maxima suppression).

\new{Our method is based on the approach of \cite{osco2020convolutional} with several modifications to improve the results.   \cite{osco2020convolutional} employ a network using a VGG-16 backbone \citep{simonyan2014very} followed by several convolutional blocks with residual connections.   They apply peak finding to the confidence map but use a fixed detection threshold without hyperparameter tuning.  Our method improves upon the method of \cite{osco2020convolutional} by replacing their network architecture with an attention-based architecture and using hyperparameter tuning to optimize the detection results.  In our evaluation comparing our method with several previous methods, we found that our proposed method produced the best results across several metrics.}

\subsection{Network Architecture}

\begin{figure*}
    \centering
    \includegraphics[width=\textwidth]{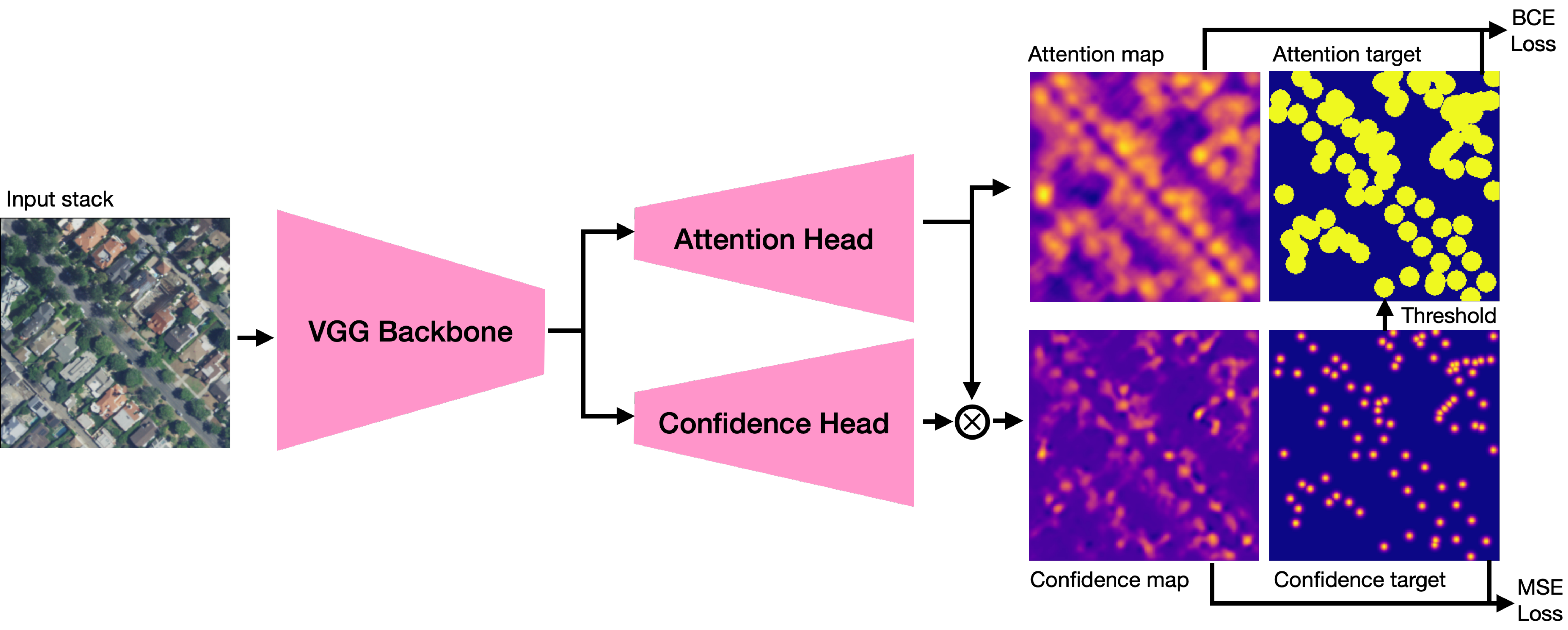}
    \caption{HR-SFANet network architecture.  The input image is encoded through the first five blocks of convolutional and pooling layers from the VGG-16 network.  Separate attention and confidence heads upsample and aggregate the outputs of the backbone layers to produce an attention map and a confidence map.  These are multiplied together to produce the final confidence map.} 
    \label{fig:cnn}
\end{figure*}

We replace the CNN architecture used previously by \cite{osco2020convolutional} with a network based on SFANet \citep{zhu2019dual}, an architecture that performed well on several crowd counting benchmarks.  Our architecture is illustrated in Figure \ref{fig:cnn}.  SFANet consists of a VGG-16 backbone \citep{simonyan2014very}, a confidence head, and an attention head.  \new{The VGG-16 backbone consists of a series of convolutions and pooling operations to encode the input, while the output heads consist of convolutions, upsampling operations, and skip connections to decode the outputs.}  We modified the SFANet architecture by adding extra layers to the confidence and attention heads so that they would output at full-resolution rather than half-resolution.  We call our network the HR-SFAnet.  The architecture is illustrated in Figure \ref{fig:cnn} \new{and documented in detail in \ref{sec:pseudocode}.}

\subsection{Data preparation}

We use NAIP multi-spectral digital number (DN) data as input, which has red (R), green (G), blue (B), and near infrared (N) channels.  We also added an NDVI channel \citep{kriegler1969preprocessing,rouse1974monitoring} derived from the red and near-infrared channels as follows:
\begin{equation}
    V = \frac{N-R}{N+R}
\end{equation}
The NDVI index is a well-known indicator of live green vegetation, and we hypothesized its inclusion would help the network identify trees in the imagery.  

Each band of the input is normalized before being processed by the network.  The NAIP multispectral DN data is eight-bit and has a range of 0-255.  The RGB bands are zero-centered with respect to the ImageNet dataset \citep{deng2009imagenet} on which the VGG16 backbone network was pre-trained.  The N band is normalized by subtracting 127.5.  The V band has a range of -1 to 1 and so we multiply by 127.5 to bring it into a similar range as the other bands after normalization.


\subsection{Network initialization}

The backbone portion of the HR-SFANet network uses a VGG-16 network \citep{simonyan2014very} with weights pre-trained on ImageNet \citep{deng2009imagenet}.  Since ImageNet contains RGB images, the VGG-16 network is not designed for inputs containing the extra near-IR and NDVI channels present in our imagery.  We remedied this by adding two extra input channels to the first layer of the VGG-16 network.  The filter weights for these channels are randomly initialized using ``Glorot'' uniform initialization \citep{glorot2010understanding}.

\subsection{Confidence map preparation}

During training, each input image tile has a corresponding ground truth confidence map that is used as target for the output of the network.  
Following \cite{osco2020convolutional} we form the target confidence map by placing a Gaussian on each tree location and aggregating them with a per-pixel maximum:
\begin{equation}
\label{eq:max}
    C(x,y) = \max_i \exp \left( -\frac{(x-x_i)^2+(y-y_i)^2}{2\sigma^2} \right)
\end{equation}
where $(x,y)$ is the location of a pixel in the confidence map and $(x_i,y_i)$ is the location of the $i$-th tree in the image.
Using the $\max$ operator (instead of a sum \citep{lempitsky2010learning}) ensures that objects have distinct peaks in the confidence map, even if they are close to each other.  

While we would ideally adapt the Gaussian width parameter $\sigma$ to each tree based on its size, our point annotations do not contain size information.  Therefore we choose a single fixed value for $\sigma$ for all trees.  We tested several settings for $\sigma$ and found $\sigma=1.8m$ to be the best setting in our experiments.

\subsection{Loss functions}

Our primary loss function is the mean squared error (MSE) loss between the predicted and ground truth confidence maps:
\begin{equation}
    L_{\textrm{MSE}} = \frac{1}{N}\sum_{x,y} [C'(x,y)-C(x,y)]^2
\end{equation}
where $C$ and $C'$ are the ground truth and predicted confidence maps, respectively.

Unlike the CNN used by \cite{osco2020convolutional}, our HR-SFANet has a an attention head in addition to the confidence head.  Following \cite{zhu2019dual}, we apply the binary cross-entropy (BCE) loss function to the output of the attention head:
\begin{equation}
\begin{split}
    L_{\textrm{BCE}} = -\frac{1}{N}\sum_{x,y}[A(x,y)\log(A'(x,y)) \\
    + (1-A(x,y))\log(1-A'(x,y))]
\end{split}
\end{equation}
where $A$ is a binary mask produced by applying a threshold $\tau$ to the ground truth confidence map, and $A'$ is the output of the attention head.  

The two loss functions are combined in a weighted sum:
\begin{equation}
    L = L_{\textrm{MSE}} + \alpha L_{\textrm{BCE}}
\end{equation}
where $\alpha$ balances the two loss terms.  

Following \cite{zhu2019dual}, we used $\tau=0.001$ and $\alpha=0.01$. We tested different settings of $\alpha$ but found that the model's performance was not sensitive to the setting of $\alpha$.

\subsection{Peak finding and hyperparameter tuning}

During inference, we produce a confidence map from the input raster and then use peak finding to determine predicted tree locations (Figure \ref{fig:inference}).  A location in the confidence map is labeled as a peak if it is the local maximum in a region of radius $2d+1$, where $d$ is the minimum allowable distance between peaks.  We also remove local maxima below a threshold, to remove false detections.  The threshold can either be an absolute threshold $t_{\textrm{abs}}$ or a relative threshold $t_{\textrm{rel}}$ (relative to the maximum value in the confidence map).

\cite{osco2020convolutional} proposed to use an absolute threshold of $t_{\textrm{abs}}=0.2$ and a minimum distance of $d=3$.  However, we found that we could increase detector performance by finding optimal settings through hyperparameter tuning.
For each configuration of the network tested, after training the model, we used hyperparameter tuning to determine the optimal settings for $d$, $t_{\textrm{abs}}$, and $t_{\textrm{rel}}$ and to choose whether to use the absolute or relative threshold.  We the used Optuna hyperparameter optimization framework \citep{optuna_2019} and searched for the optimal set of hyperparameters over 200 iterations.  We retained the settings with the highest F-score over the validation set.  In our experiments, we found that, in contrast to the recommendation of \cite{osco2020convolutional}, the relative threshold was always the better choice.

\subsection{Implementation details}

We implemented our HR-SFANet in Python using the Tensorflow \citep{abadi2016tensorflow} and Keras \citep{chollet2015keras} libraries.  Training and inference were performed on an Nvidia V100 GPU.  

During training, we processed batches of $256 \times 256$ tiles from the training set, using a batch size of eight over 500 epochs.  We used the Adam optimizer \citep{kingma2014adam} with learning rate $0.0001$, $\beta_1=0.9$, $\beta_2=0.999$, and $\epsilon=1 \times 10^{-7}$.  We reserved 10\% of the training data for validation and retained the model with best validation loss during training.  For data augmentation, we rotated each training patch by 90$^\circ$, 180$^\circ$, and 270$^\circ$ and also horizontally flipped each patch.  This increased the size of the training set eight-fold.  We experimented with further data augmentation using random image rotation and brightness/color variation but found that it did not lead to an improvement.

\subsection{Other tree detection methods}

We also tested two other well-known tree detection methods in our evaluation: PyCrown \citep{zorner2018lidar} and DeepForest \citep{weinstein2019individual}.

\subsubsection{PyCrown}

PyCrown \citep{zorner2018lidar} uses LiDAR data as input and applies peak finding on a smoothed canopy height model (CHM) raster to determine tree locations, with some filtering and post-processing applied to eliminate false positives and increase localization accuracy.  The method is an open-source re-implementation of \cite{dalponte2016tree} in Python with some modifications to improve speed and produce better results.    Because PyCrown tends to produce many false positives on buildings, we removed any tree detections that landed within OpenStreetMap building extents \citep{haklay2008openstreetmap}.  Since public LiDAR data is not currently available in Palm Springs, CA, we removed those regions from the test set when evaluating PyCrown.

\subsubsection{DeepForest}

DeepForest \citep{weinstein2019individual} is a Python package that provides a pre-trained tree detection model based on the RetinaNet object detector \citep{lin2017focal}.  The pre-trained model was trained on RGB imagery from sites in the National Ecological Observatory Network (NEON).  The method outputs a bounding box for each detected tree.

\section{Experiments}
\label{sec:experiments}

\subsection{Evaluation metrics}
\label{sec:metrics}

We evaluate our models on five standard metrics: Average Precision (AP), precision, recall, F-score, and Root Mean Square Error (RMSE).  Given the count of true positives (TP), false positives (FP), and false negatives (FN), precision, recall, and F-score are calculated as follows:
\begin{eqnarray}
\textrm{precision} &= \frac{\textrm{TP}}{\textrm{TP}+\textrm{FP}} \\
\textrm{recall} &= \frac{\textrm{TP}}{\textrm{TP}+\textrm{FN}} \\
\textrm{F-score} &= 2 \frac{\textrm{precision} \cdot \textrm{recall}}{\textrm{precision} + \textrm{recall}}
\end{eqnarray}
True positives, false positives, and false negatives are determined by matching predicted tree locations and ground truth tree locations.  We find the minimum weight matching, which minimizes the sum of distances between matched trees, and ensures that each ground truth tree and each predicted tree has at most one match.  We remove matches whose distance is over a threshold; in our evaluation we used a threshold of six meters.  Example results from the matching procedure can be seen in Figure \ref{fig:crops}.

RMSE is calculated from the distance between true positive predictions and their corresponding ground truth trees.  Let $\mathcal{T}$ be the set of accepted matches and let $(x_i,y_i)$ and $(\hat{x}_i,\hat{y}_i)$ be the locations of corresponding predicted and ground truth trees, respectively.
\begin{equation}
\textrm{RMSE} = \sqrt{ \frac{1}{|\mathcal{T}|} \sum_{i \in \mathcal{T}} (x_i-\hat{x}_i)^2 + (y_i - \hat{x}_i)^2}
\end{equation}

To calculate AP, we calculate precision and recall over a range of absolute thresholds for accepting predicted trees according to their confidence value (the value of the confidence map at the tree's location).  Let $P_n$ and $R_n$ be the precision and recall observed at the $n$-th threshold.  The AP is calculated as:
\begin{equation}
\textrm{AP} = \sum_n (R_n - R_{n-1}) P_n
\end{equation}
AP is useful because it summarizes the performance of the detector across a range of detection thresholds, whereas the other metrics describe the performance of the detector at a single detection threshold optimized to balance both precision and recall.

To calculate evaluation metrics for DeepForest, we considered each ground truth point contained within a predicted bounding box as a possible true positive.  We calculated an optimal matching to determine the assignment between ground truth points and bounding boxes which maximized the number of true positives.  To calculate RMSE for DeepForest we used the center points of the bounding boxes for the predicted tree locations.

\subsection{Comparison of methods on Southern California test set}

\begin{table*}[t]
    \centering
    \caption{Comparison of tree detection methods on Southern California 2020 test set.}
    \begin{tabular}{l|rrrrr}
    \textbf{Method} & \textbf{AP} & \textbf{Precision} & \textbf{Recall} & \textbf{F-Score} & \textbf{RMSE [m]} \\
    \hline
PyCrown \cite{zorner2018lidar}
& - & 0.401 & 0.661 & 0.499 & 2.709 \\
DeepForest \cite{weinstein2019individual} 
& 0.387 & 0.735 & 0.294 & 0.420 & 2.719 \\
\hspace{1em} + fine-tuning
& 0.701 & 0.707 & 0.713 & 0.710 & 2.431 \\
\hspace{1em} + hyperparameter tuning 			          
& 0.701 & 0.782 & 0.683 & 0.729 & 2.413 \\
Osco et al. \cite{osco2020convolutional}
& 0.660 & \textbf{0.803} & 0.476 & 0.598 & 2.270 \\
\hspace{1em} + hyperparameter tuning
& 0.660 & 0.764 & 0.706 & 0.734 & 2.263 \\
HR-SFANet (ours)	  
& \textbf{0.705} & 0.736 & \textbf{0.733} & \textbf{0.735} & \textbf{2.157} \\
    \end{tabular}
    \label{tab:method_comparison}
\end{table*}

\begin{table*}[t]
    \centering
    \caption{Results of our method when extrapolating to different climate zones and years.  The top portion consists of the same Southern California cities in the 2020 training set, but earlier image capture years, while the bottom portion consists of Northern California cities distinct from the training set region.}
    \begin{tabular}{lrl|rrrrr}
    \textbf{City} & \textbf{Year} & \textbf{Zone} & \textbf{AP} & \textbf{Precision} & \textbf{Recall} & \textbf{F-Score} & \textbf{RMSE [m]} \\
    \hline
Santa Monica &  2016 & SCC & 0.665 &      0.734 &   0.693 &    0.713 &     2.132 \\
Santa Monica &  2018 & SCC & 0.707 &      0.732 &   0.730 &    0.731 &     2.188 \\
  Long Beach &  2016 & SCC & 0.694 &      0.799 &   0.668 &    0.728 &     1.963 \\
  Long Beach &  2018 & SCC & 0.720 &      0.856 &   0.639 &    0.732 &     1.830 \\
   Claremont &  2016 & IE & 0.654 &      0.739 &   0.668 &    0.701 &     2.075 \\
   Claremont &  2018 &  IE & 0.642 &      0.708 &   0.668 &    0.687 &     2.435 \\
   Riverside &  2016 &  IE & 0.723 &      0.819 &   0.671 &    0.737 &     1.905 \\
   Riverside &  2018 &  IE & 0.590 &      0.686 &   0.622 &    0.652 &     2.599 \\
Palm Springs &  2016 &  SD & 0.624 &      0.700 &   0.647 &    0.672 &     1.953 \\
Palm Springs &  2018 &   SD & 0.645 &      0.743 &   0.620 &    0.676 &     1.804 \\
\hline
       Chico &  2018 &   IV & 0.708 &      0.748 &   0.688 &    0.716 &     2.179 \\
       Chico &  2020 &  IV & 0.701 &      0.733 &   0.689 &    0.710 &     2.338 \\
      Eureka &  2020 & NCC & 0.557 &      0.744 &   0.509 &    0.604 &     2.423 \\
      Bishop &  2020 &  IW & 0.687 &      0.723 &   0.694 &    0.708 &     2.203 \\
    \end{tabular}
    \label{tab:extrapolation_results}
\end{table*}

We tested each method on the Southern California 2020 portion of the dataset.  The results are summarized in Table \ref{tab:method_comparison}.

\subsubsection{Our method}
 
As our method is based on \cite{osco2020convolutional} we first tested their method on our dataset.  Since their code is not available, we created our own implementation based on the description in the paper and trained the network from \cite{osco2020convolutional} in the same manner as our method.  Using their recommended settings for peak finding, the Osco et al.~method achieves an AP of 0.660, F-Score of 0.598, and RMSE of 2.270 m.  After applying our proposed method of hyperparameter tuning to find the optimal peak finding settings, F-Score increases to 0.734, and RMSE decreases to 2.263 m.    When we replace their network with our HR-SFANet, AP increases to 0.705, F-Score increases to 0.735, and RMSE decreases to 2.157 m.  This experiment validates the importance of our proposed network design and hyperparameter tuning approach to improve the results.

\subsubsection{PyCrown}
PyCrown only achieved an F-score of 0.499 and RMSE of 2.709 m on our test set.  We could not calculate AP for PyCrown because there is no detection threshold to vary in this method.  The precision for PyCrown was exceptionally low (0.401), indicating that it output more false positives than other methods.  This could be due to a single tree having multiple local peaks in its canopy height map, leading to false detections.

\subsubsection{DeepForest}

The pre-trained DeepForest model achieved a high precision (0.735) but low recall (0.294) , indicating that it was unable to detect many trees that other methods could detect.  This led to an F-Score of 0.420 and AP of 0.387.  This poor performance is likely due to the fact that the NEON imagery on which DeepForest was trained is higher resolution than our NAIP imagery, and mostly contains wilderness areas whereas our imagery contains urban environments.

To improve the results of DeepForest, we fine-tuned the pre-trained model on our training set.  Because our data does not include bounding box annotations, we placed a fixed size bounding box around each ground truth point.  We experimented with different box sizes and chose the size that led to the best results on the test set, which was a box with width and height of 9.6 m (16 pixels).   We fine-tuned the model for 15 epochs; training past this point did not yield any further improvement.  After fine-tuning, the AP increased substantially to 0.701, and the F-Score increased to 0.710 using the default detection threshold setting.   After hyper-parameter tuning, the F-Score improved to 0.729 with an RMSE of 2.413.  This experiment indicates that a bounding box object detector can be trained to effectively detect trees using point annotations; however, the performance is below what we achieved with our confidence map approach.

\subsubsection{Inference speed}

We calculated the average processing time for each method to detect trees in a $256\times256$ raster.  PyCrown is the slowest method, taking over five seconds per image.  The CNN-based methods are much faster, and ours is the fastest.  DeepForest takes 71 ms per image, \cite{osco2020convolutional} takes 44 ms per image, and our method takes 21 ms per image.

\subsection{Extrapolation to other years and climate zones}

\begin{figure*}
    \centering
    \begin{subfigure}[b]{.25\textwidth}
         \includegraphics[width=\textwidth]
         {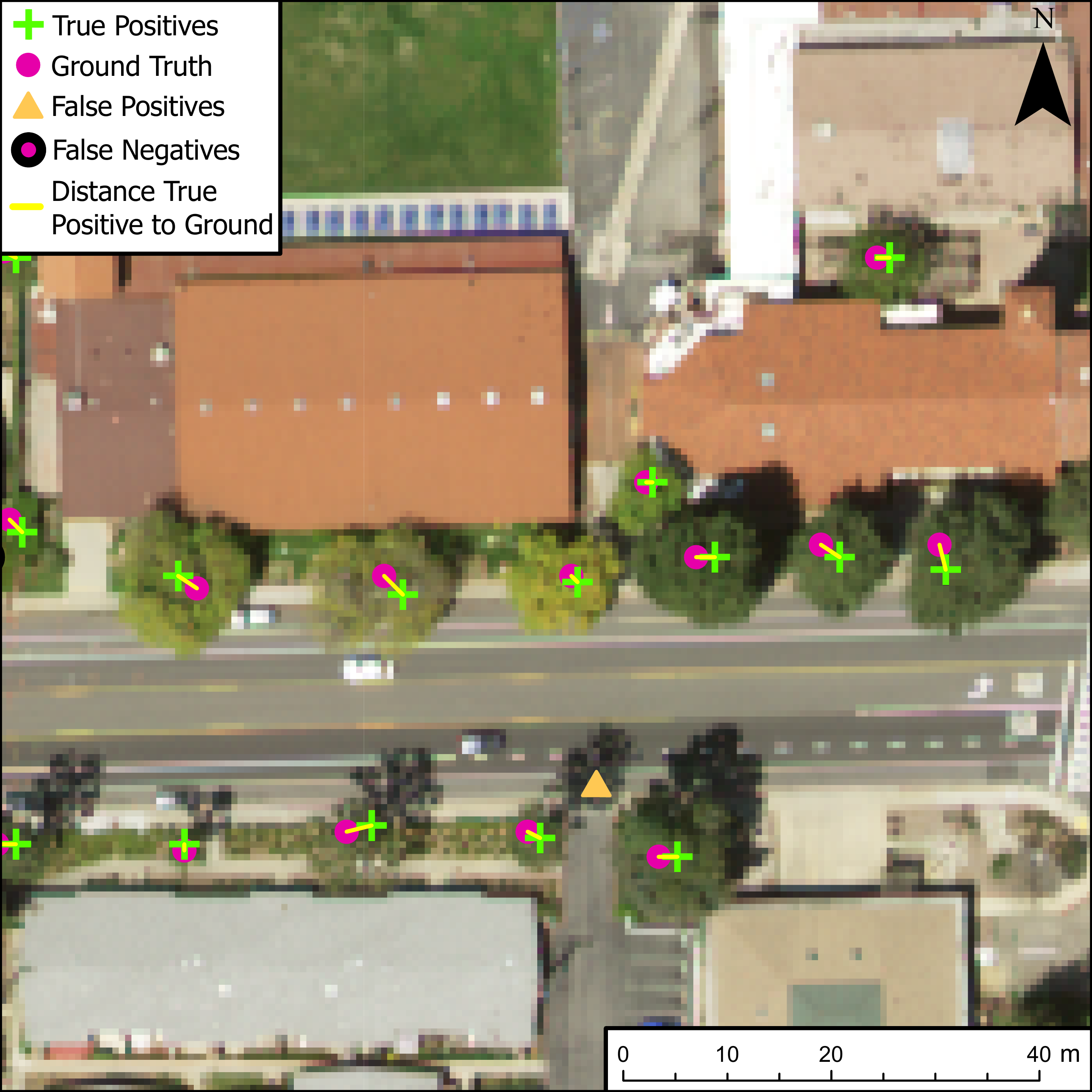}
         \caption{\label{subfig:claremont_good}Claremont}
    \end{subfigure}
    \begin{subfigure}[b]{.25\textwidth}
         \includegraphics[width=\textwidth]
         {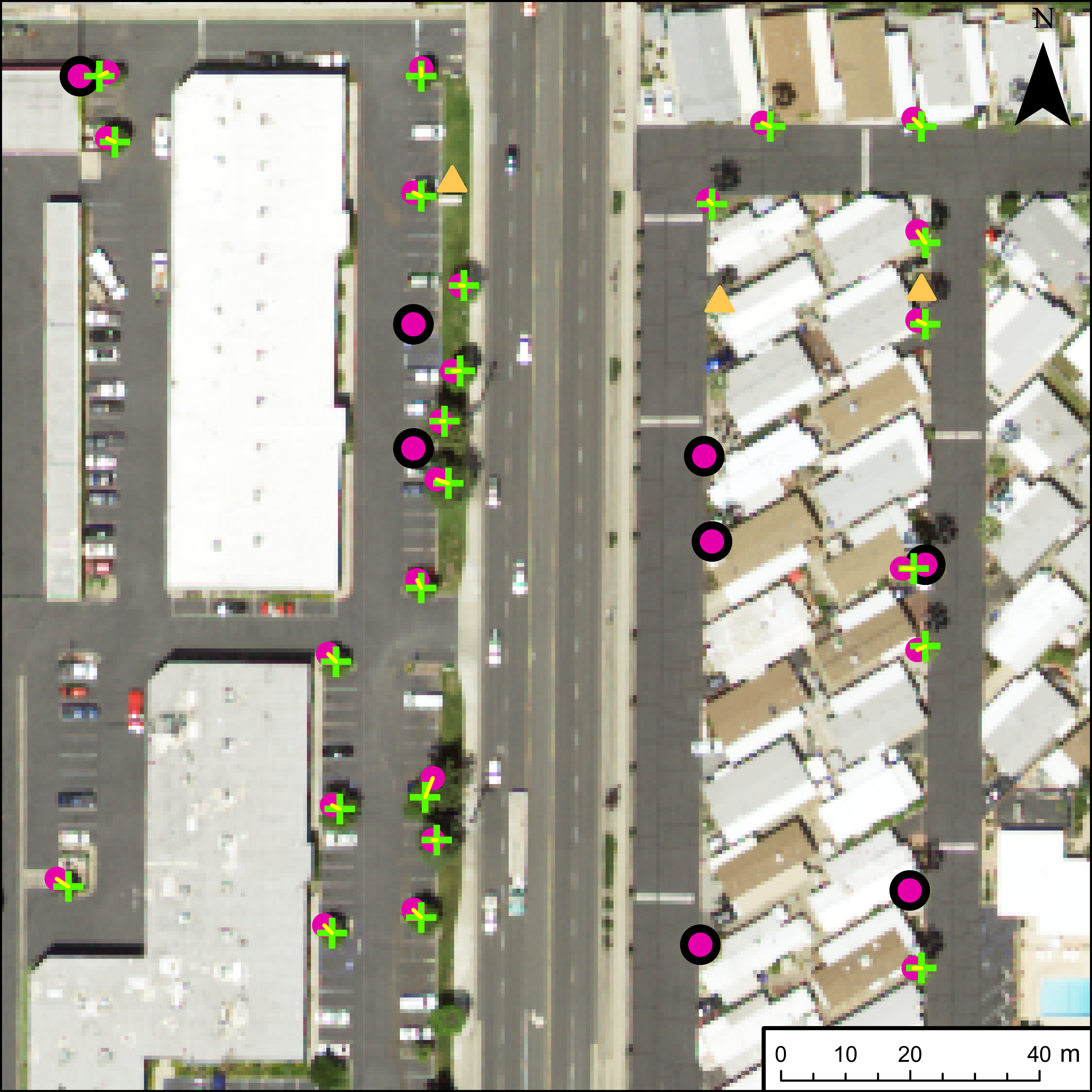}
         \caption{\label{subfig:claremont_industrial_good_no_legend}Claremont}
    \end{subfigure}
    \begin{subfigure}[b]{.25\textwidth}
         \includegraphics[width=\textwidth]
         {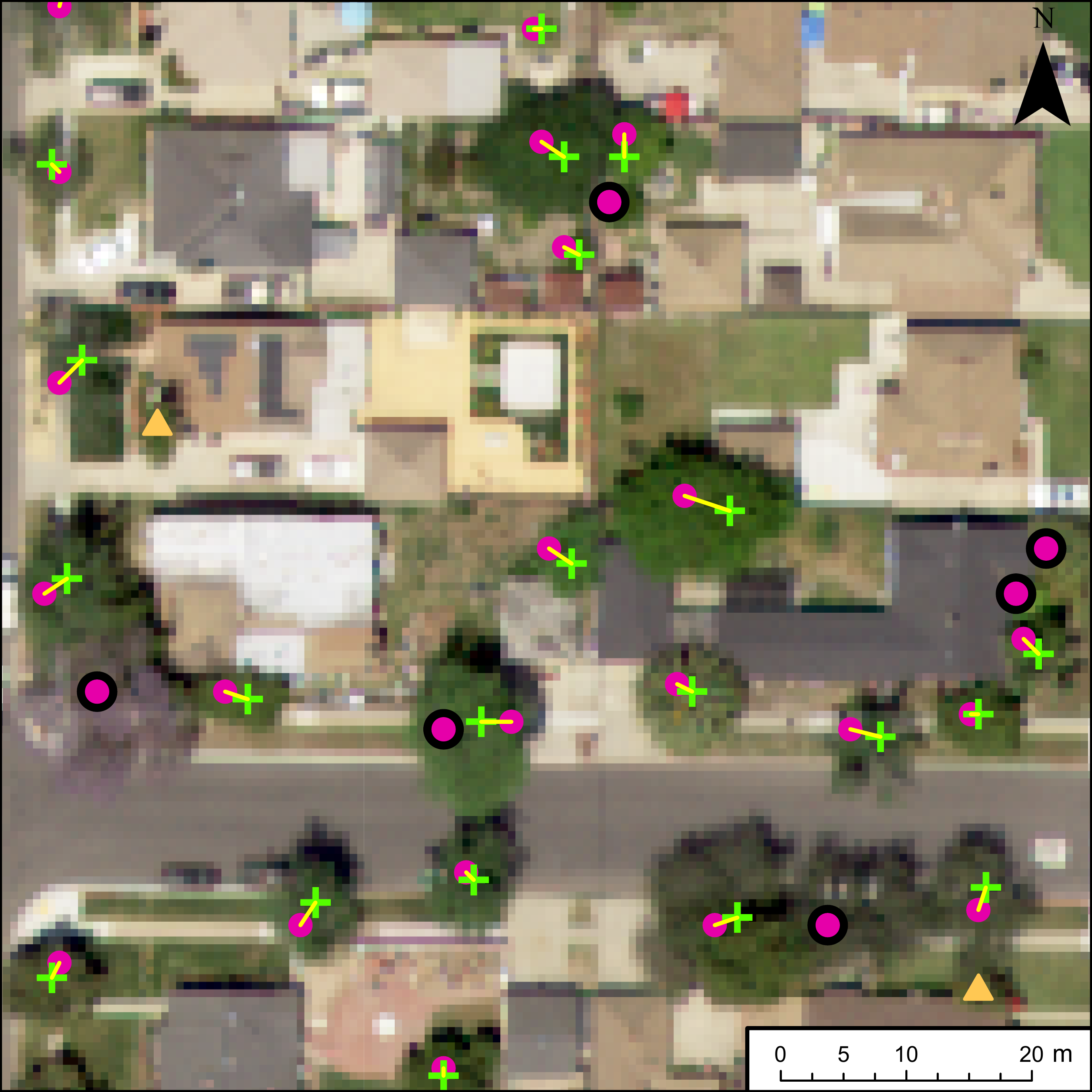}
         \caption{\label{subfig:long_beach_false_negatives}Long Beach}
    \end{subfigure}\\
    
    \begin{subfigure}[b]{.25\textwidth}
         \includegraphics[width=\textwidth]
         {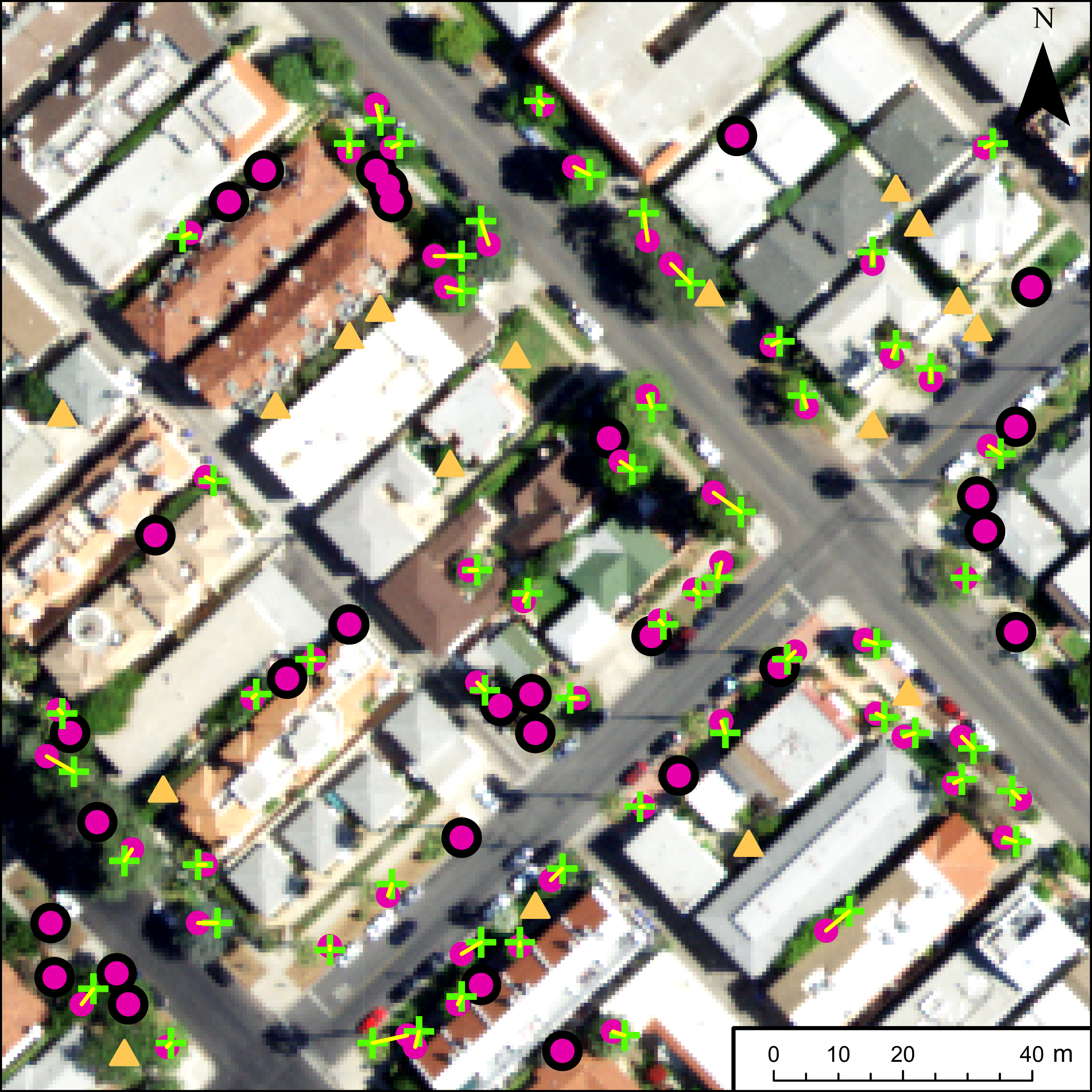}
         \caption{\label{subfig:santa_monica_false_negatives}Santa Monica}
    \end{subfigure}
    \begin{subfigure}[b]{.25\textwidth}
         \includegraphics[width=\textwidth]
         {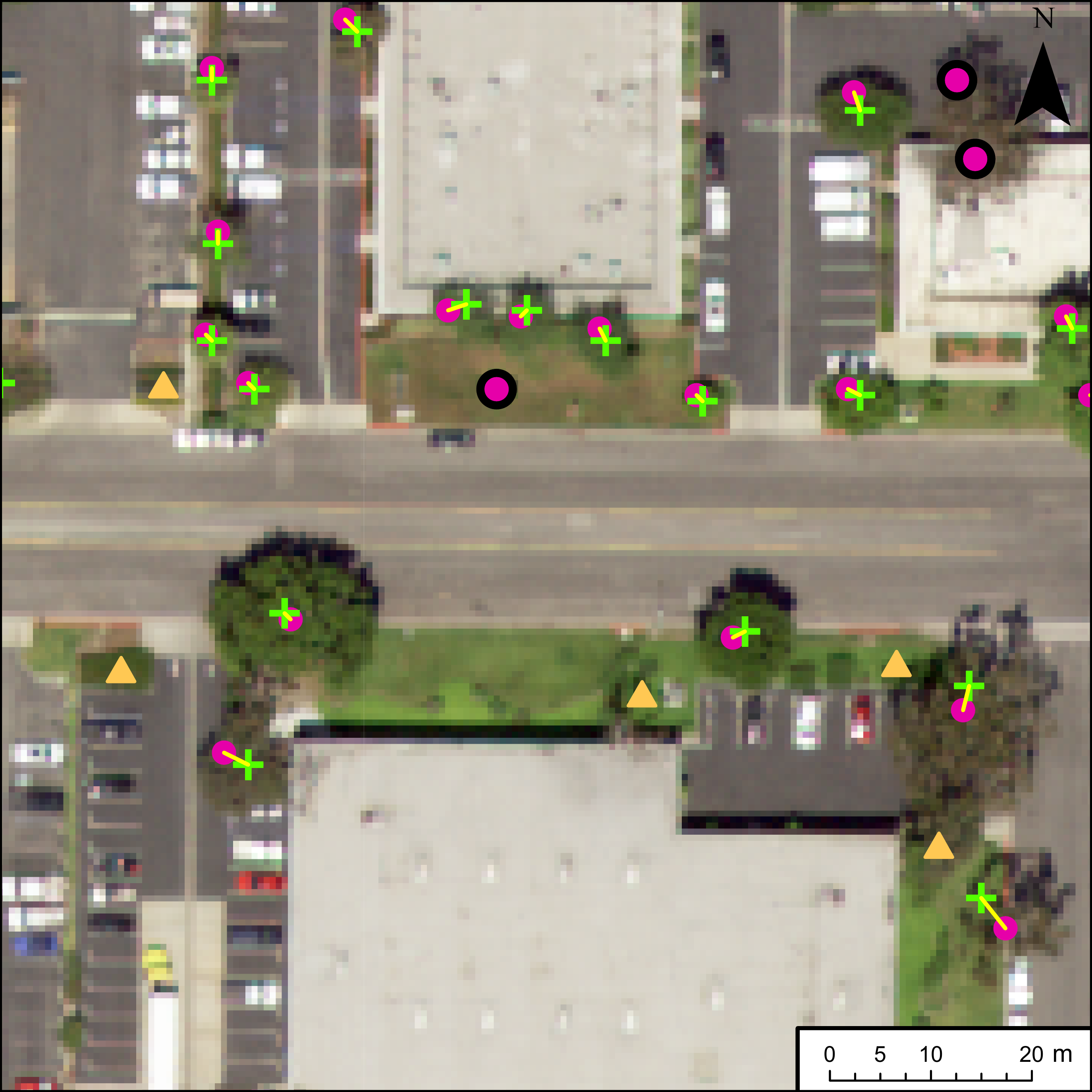}
         \caption{\label{subfig:long_beach_false_positives}Long Beach}
    \end{subfigure}
    \begin{subfigure}[b]{.25\textwidth}
         \includegraphics[width=\textwidth]
         {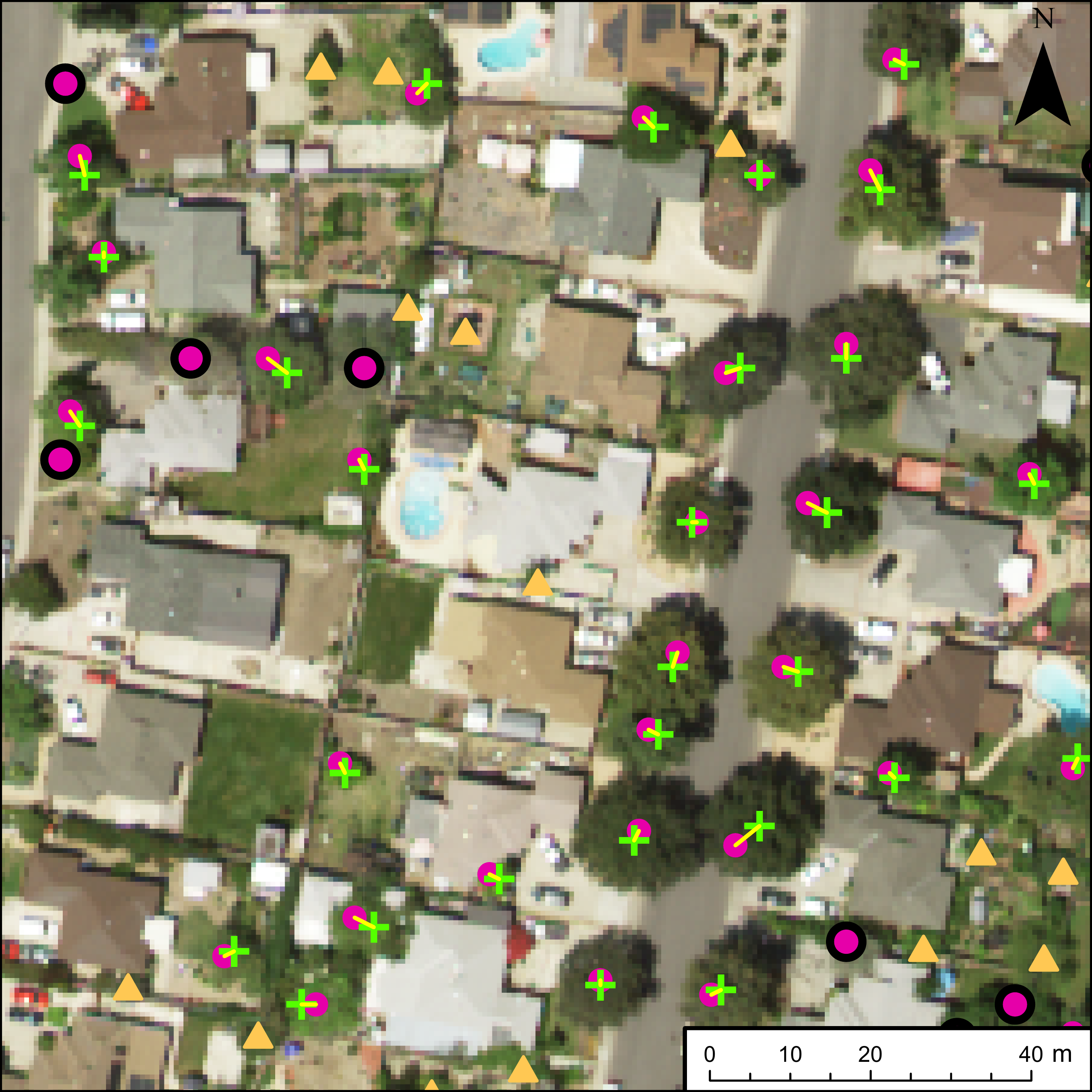}
         \caption{\label{subfig:claremont_false_positives}Claremont}
    \end{subfigure}
    
    \caption{Example results from our tree detection method on images from Southern California 2020 test set.  The method is able to detect and accurately localize most of the trees in the images.  Some areas contain examples of missed detections of trees with small canopies and shrubs being confused with trees.}
    \label{fig:crops}
\end{figure*}

When trained and tested on 2020 imagery from Southern California, our method achieved the best results among methods tested.  We then evaluated the performance of our method on pre-2020 Southern California imagery and on imagery from three cities in Northern California, each in a different climate zone.  None of the images in these experiments were present in the training or validation data.  Because these test sets represent different climate zones, varying tree density, and in some cases an earlier image capture year than the training set (see Table \ref{tab:dataset_summary}), they serve to evaluate our method's ability to generalize to data different from the training set.  The results are summarized in Table \ref{tab:extrapolation_results}.  Overall, the precision in these areas was similar to the precision observed in the 2020 Southern California test set, ranging from 0.856 to 0.686, but recall was lower, ranging from 0.730 to 0.509.  The F-score ranged from 0.737 to 0.604.

\begin{figure}[t]
    \centering
    \includegraphics[width=\columnwidth]{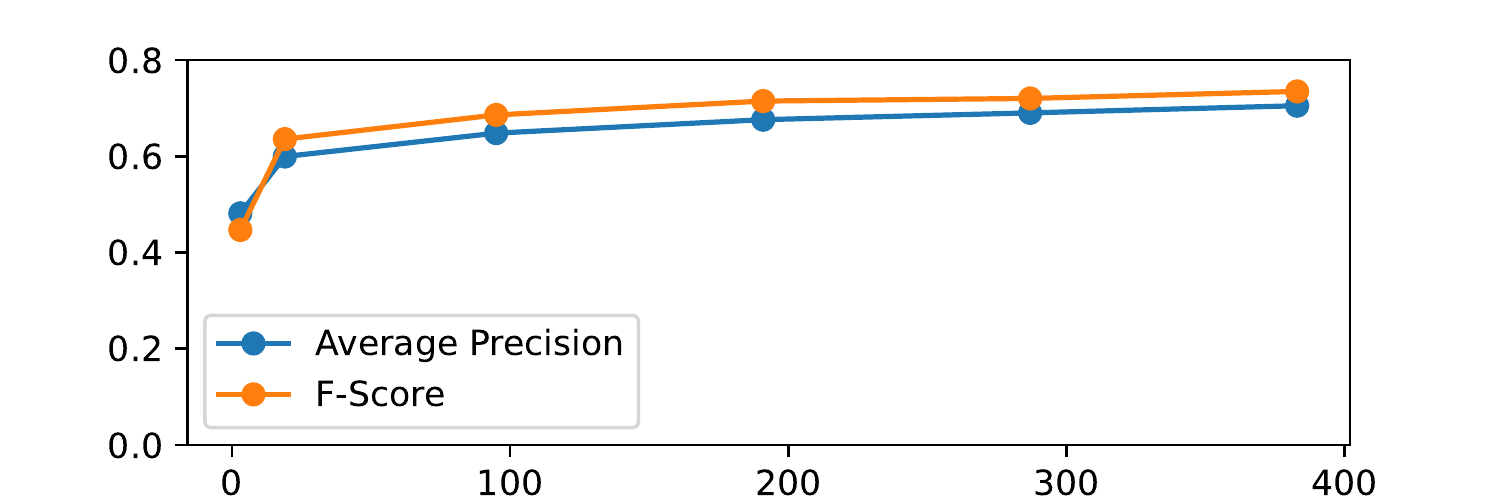}
    \caption{Analysis of test set performance with increasing training set size.}
    \label{fig:training_data_size}
\end{figure}

\subsection{Qualitative analysis}

Figure \ref{fig:crops} shows example results from our method on selected regions from our test sets.  In general, we noticed three typical sources of errors: missed detections of trees with small canopies; confusion of different types of plant, such as shrubs, with trees; and over- or under-estimation of trees in areas of dense canopy, which sometimes have strong shadows. Another challenge is the variation in viewing angle, illumination, and canopy size seen in the NAIP imagery, since the time and date of capture varied across regions and from year to year.  These sources of errors would be mitigated by adding more training data or fine-tuning on specific regions, and using imagery with higher spatial and spectral resolution.

\subsection{Effect of reducing training set size}

\begin{table}[t]
    \centering
    \caption{Test set results for our HR-SFANet model trained on 90\% of the entire dataset, covering all years and climate zones.  The test set is the remaining 10\% of the dataset.}
    \begin{tabular}{rrrrr}
    \textbf{AP} & \textbf{Precision} & \textbf{Recall} & \textbf{F-Score} & \textbf{RMSE [m]} \\
    \hline
    0.727 & 0.765 & 0.720 & 0.741 & 1.936 
    \end{tabular}
    \label{tab:retrain}
\end{table}

Deep learning methods are well-known for requiring large amounts of training data to be effective; however, manual annotation of images is time-intensive and tedious work.  We used the Southern California 2020 portion of our dataset to investigate the relationship between training set size and test set performance for our task.  We subsampled the training set to various amounts, trained a separate network with each subset, and calculated the performance metrics for each network using the complete test set.  The results are shown in Figure \ref{fig:training_data_size}.  With 1\% of the data (3 images), the F-Score is reduced to 0.447 and AP is 0.481.  With 5\% of the data (19 images), F-Score improves to 0.636 and AP to 0.600.  From there, the metrics more slowly increase with increasing training set size.  This experiment shows that there are diminishing gains in adding more training data.

\subsection{Effect of increasing training set size}

\begin{figure}
    \centering
    \begin{subfigure}[b]{.225\textwidth}
         \includegraphics[width=\textwidth]
         {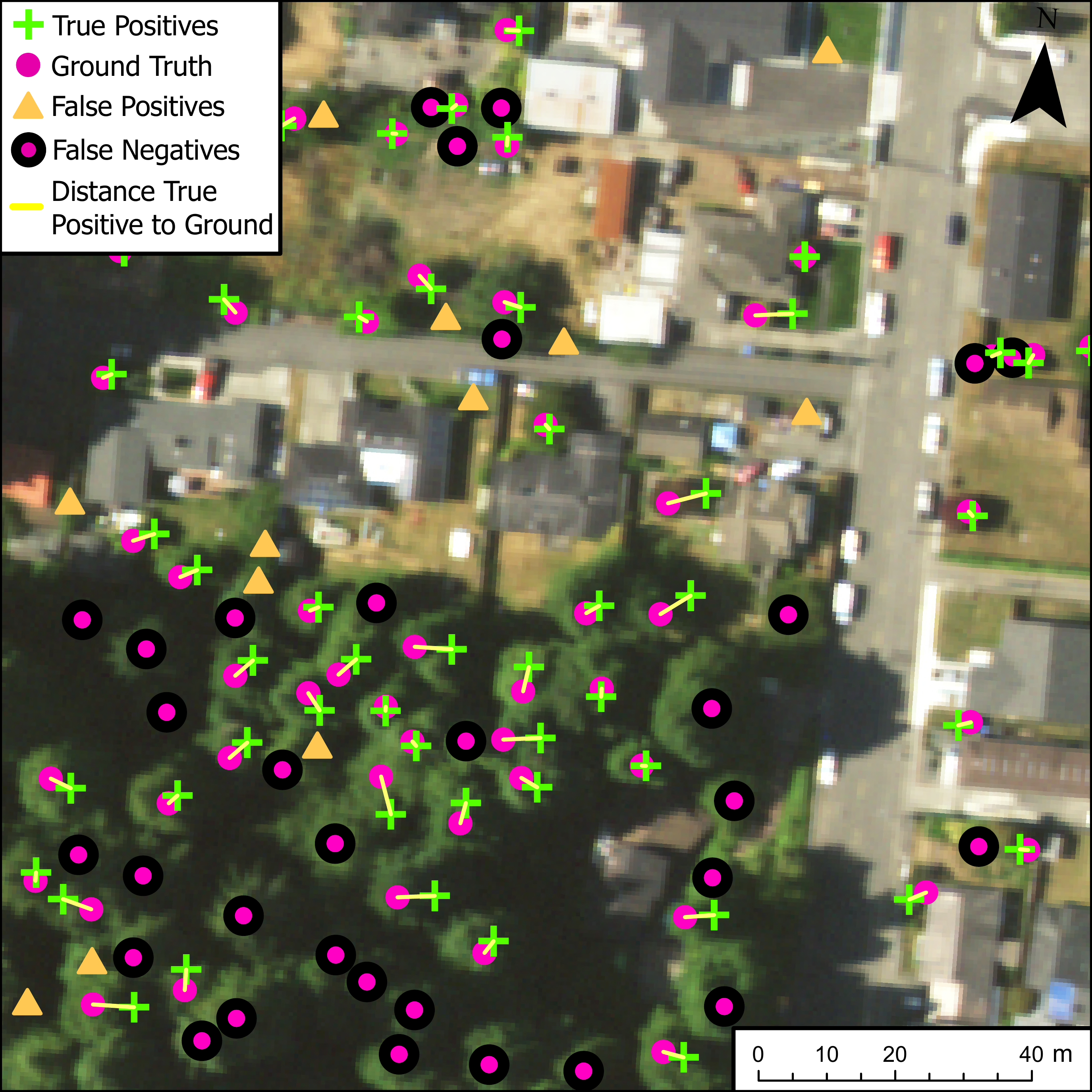}
         \caption{\label{subfig:eureka_old_model}Eureka (Original)}
    \end{subfigure}
    \begin{subfigure}[b]{.225\textwidth}
         \includegraphics[width=\textwidth]
         {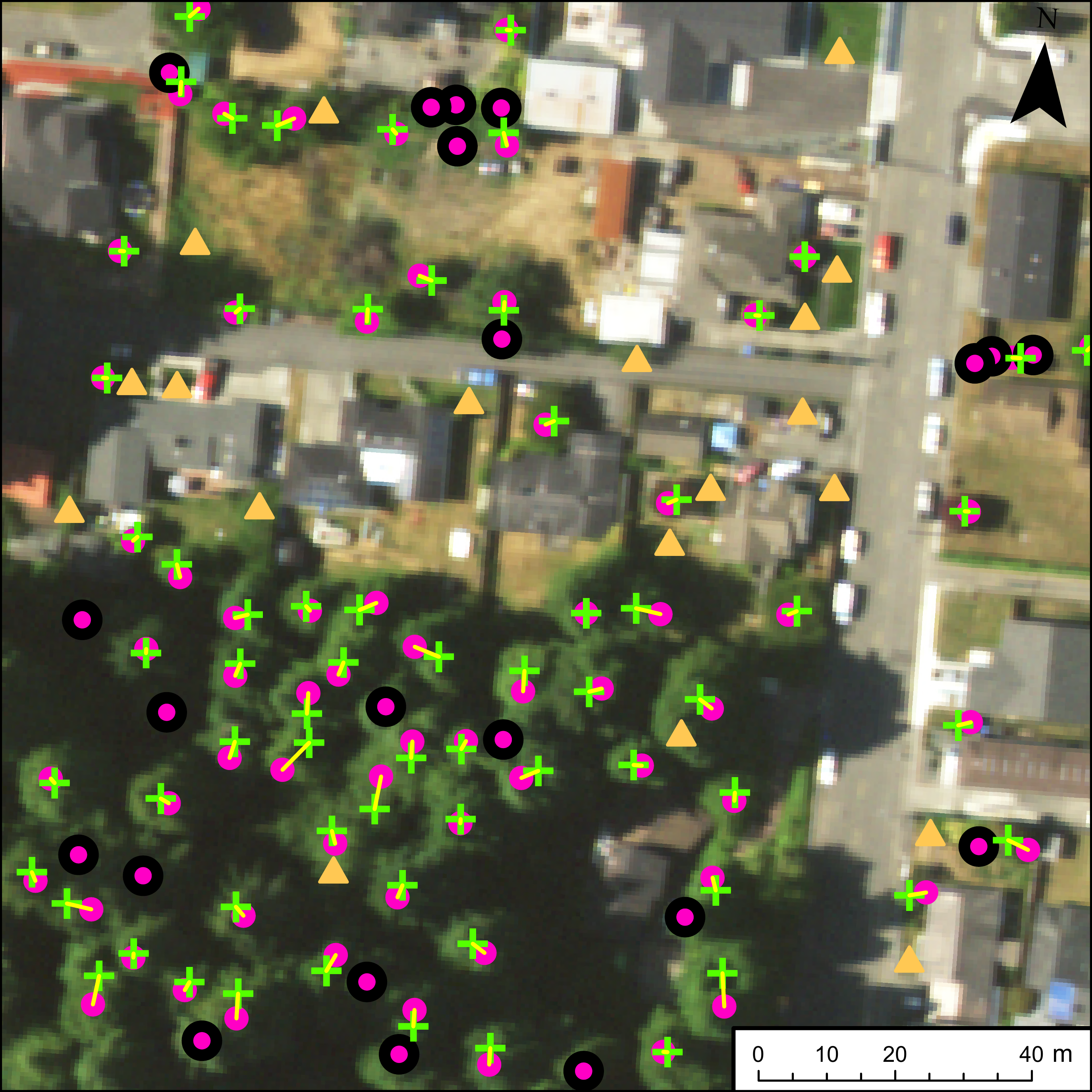}
         \caption{\label{subfig:eureka_new_model}Eureka (Re-trained)}
    \end{subfigure}    
    \caption{Results in Eureka (a) before and (b) after re-training model with data from all areas and image years.  After re-training, the model detects more trees in the dense and shadowed areas.}
    \label{fig:new_model}
\end{figure}

The lowest recall (0.509) was observed in Eureka, a city in Northern California which contains dense natural forest areas within its urban boundary (Figure \ref{fig:new_model}).   To evaluate the potential to address sources of error and improve performance by adding more training data, we re-trained the model using a random 90\% subset of the entire dataset \new{(1,485 images)}, not just the Southern California 2020 portion as in the original model.  The results of the original and re-trained model in Eureka are compared in Figure \ref{fig:new_model}, which illustrates how the re-trained model detects trees in the dense and shadowed region that were missed by the original model.  Note that the test image shown in Figure \ref{fig:new_model} was not in the training set of either model.

\new{Quantitative results for the re-trained model are shown in Table \ref{tab:retrain}.   The re-trained model improved AP to 0.727, F-score to 0.741, and RMSE to 1.936 m.  However, note that these numbers are not directly comparable to the results for the models trained and tested on the Southern California 2020 subset (Table \ref{tab:method_comparison}), because the test sets for the two evaluations are not the same.}

\begin{figure*}[t]
    \centering
     \begin{subfigure}[b]{0.32\textwidth}
         \centering
    \includegraphics[width=\textwidth]{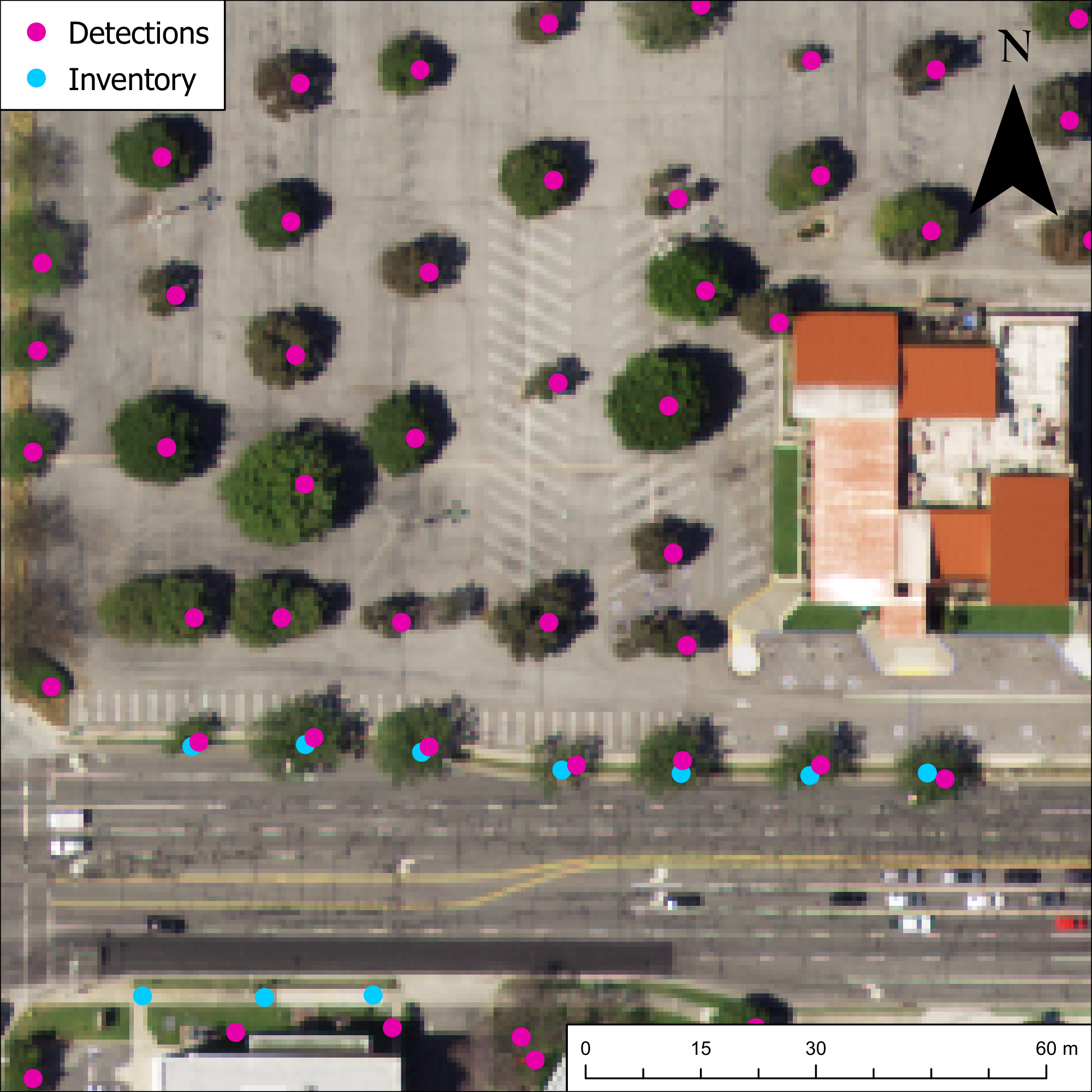}
     \end{subfigure}
     \hfill
     \begin{subfigure}[b]{0.32\textwidth}
        \centering
        \includegraphics[width=\textwidth]{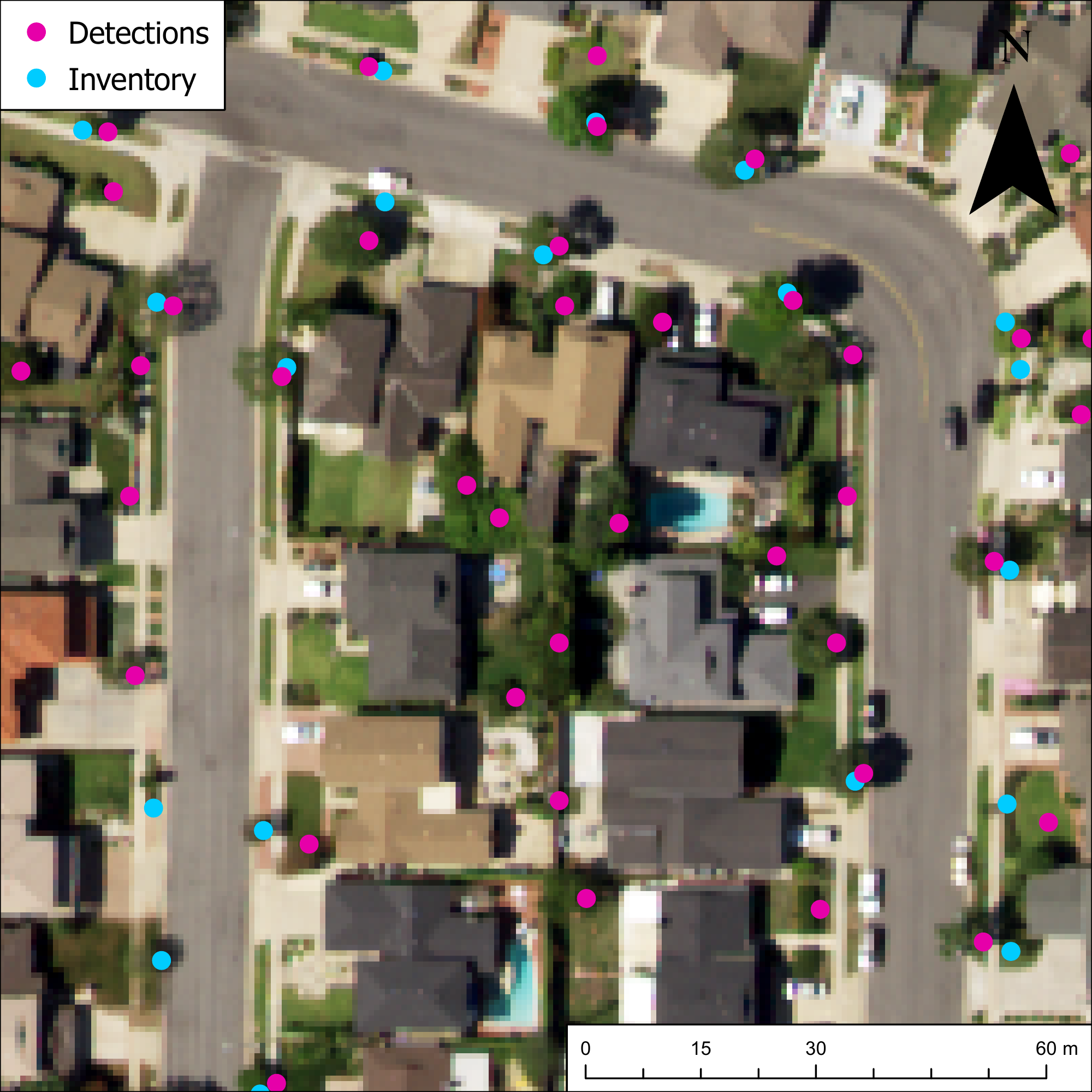}
    \end{subfigure}
    \hfill
    \begin{subfigure}[b]{0.32\textwidth}
        \centering
        \includegraphics[width=\textwidth]{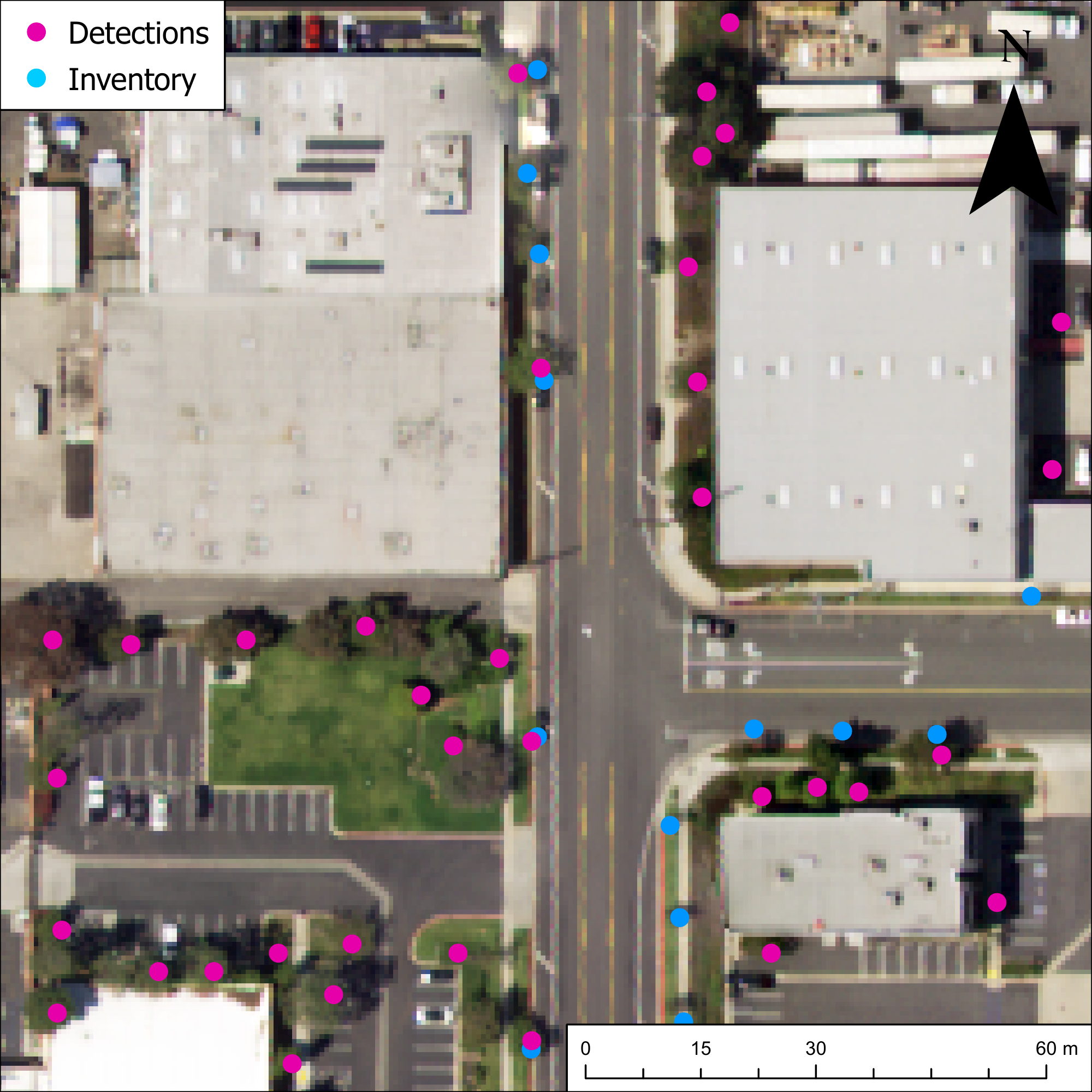}
    \end{subfigure}
    \caption{Comparison of a manual tree inventory with our tree detection results on three areas from Torrance, CA.  Our tree detector is able to find trees on both public and private land, leading to a more complete inventory.}
    \label{fig:torrance}
\end{figure*}

\section{Application}
\label{sec:application}


Because the HR-SFANet network in our method is fully convolutional \citep{long2015fully}, it can process any size raster as input.  However, a large raster will need to be processed in tiles to stay within the memory limitations of the machine.  The na\"{i}ve approach of processing each tile without overlap between the tiles results in disagreement at the edges of the confidence maps, leading to inaccurate, missing, or duplicated tree detections.  To avoid these artifacts when processing a large raster, we divide the raster into an overlapping grid of tiles.  After processing by the network, the overlap region is discarded to avoid edge artifacts.  We used a tile size of $2112 \times 2112$ pixels with an overlap of $32$ pixels.  We similarly apply local peak finding on an overlapping grid when processing a large raster; for peak finding we used a tile size of $256 \times 256$ and an overlap of $32$ pixels.

Using our method with this tiled inference approach, we processed NAIP 2020 imagery of all of California's urban areas.  Because our focus is on creating tree inventories for cities, we excluded any tree detections outside of urban boundaries as determined by the California Department of Water Resources \citep{urbanboundaries}

Our automatic tree detection approach can fill deficiencies in tree inventories that typically only cover street trees on public rights-of-way areas and thus miss a large proportion of the trees in a city's urban forest. 
For example, Figure \ref{fig:torrance} compares a manual tree inventory from Torrance, CA \citep{love2022diversity} with our automatic tree detection results. The blue points are from the existing urban forest inventory, and the purple points are trees detected with our method.  Our tree detector is able to detect trees in private spaces that are not labeled in the public inventory.

\begin{figure*}[t]
    \centering
    \includegraphics[width=\textwidth]{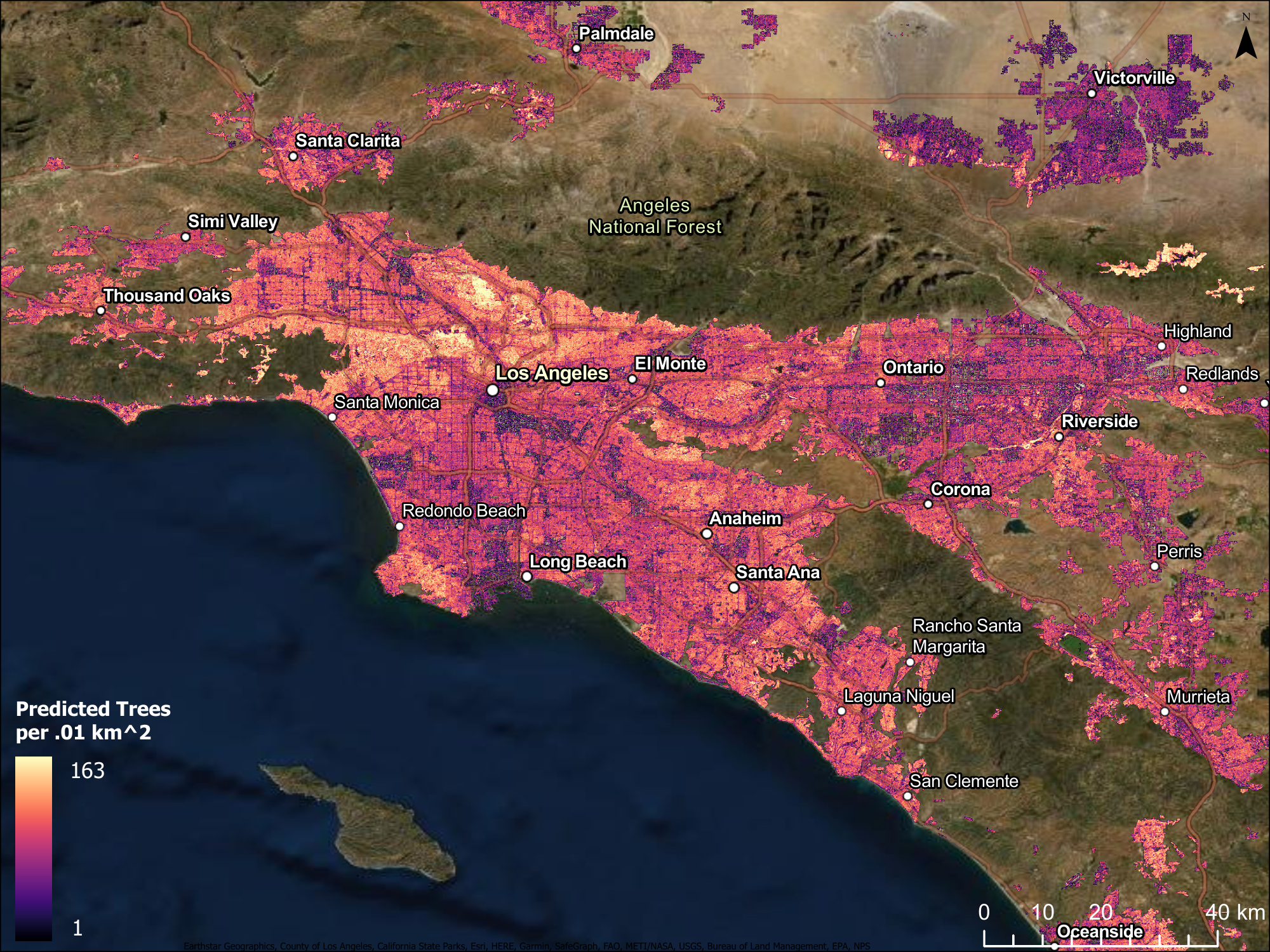}
    \caption{Visualization of tree count per .01 km$^2$ in the Los Angeles area. Each .01 km$^2$ is visualized using the total number of trees we detected in that grid after applying the Tree Detector to 2020 NAIP imagery. The basemap shown is is the World Imagery Basemap accessed through ArcGIS Pro \cite{ESRI_2023}. }
    \label{fig:heatmap}
\end{figure*}

Our method can also support larger-scale analyses of tree density across the state.  For example, Figure \ref{fig:heatmap} shows a tree density map for urban areas in Southern California. To create this map, we calculated the number of detected trees per grid cell over a 100 m $\times$ 100 m grid.

\section{Conclusions and future work}
\label{sec:conclusions}

Creating good data and efficient systems for the management of our urban forests will be essential as our climate changes, and the ecosystem services, such as controlling microclimate, of our urban forests become increasingly important \citep{mcpherson2016structure}. Currently, inventories of a city's urban forest are completed manually, and repeated on a routine basis \citep{nielsen2014review}.  This time-consuming process costs cities significant time and financial resources, and typically only accounts for the trees within the publicly managed urban forest. Inventories are used by cities to manage their urban forest, and used by cities and researchers to estimate the ecosystem services given by their forests \citep{mcpherson2016structure,love2022diversity}. Trees on privately managed land, like public trees, contribute to the ecosystem services an urban forest provides to its residents, but there are few ways to account for those services in typical tree inventories and resulting research. Our research allows for trees in the entire urban area to be counted, creating more accurate total tree estimates for cities and neighborhoods. Because our counts are spatially explicit, managers can use them to identify areas with relatively few urban trees, and attempt to address why that is the case and target that area for planting. Our data can be used to compare socioeconomic trends with tree counts, find trends in where there are more publicly or privately managed trees, and identify areas of tree inequity in urban areas.  By comparing tree inventories across years, we could analyze the progress of the urban forest over time and better understand the effect of environmental conditions such as drought.


Future work could include exploring the use of imagery with higher spatial and spectral resolution, to help the detector separate out individual tree canopies in dense stands of trees, detect small trees, and distinguish trees from other visually similar plants.  It also will be important to consider how to easily enable end users to fine-tune the model for their particular region of interest.

Furthermore, while our method accurately detects and localizes trees from aerial imagery, on-the-ground urban tree inventories provide much more information than our current method can provide, such as: coarse-grained classification of the tree (e.g., identifying deciduous and coniferous trees); fine-grained classification of the tree genus or species; estimation of canopy size; health status of the tree; and monitoring of silvicultural treatments.  The ability to produce such information would be highly valuable for improving an urban tree inventory.  Future work lies in incorporating techniques such as image-based classification \citep{beery2022auto} and canopy size estimation \new{using instance segmentation} \citep{sun2022counting} to automatically determine these important tree properties from imagery.

\section{Acknowledgments}

This project was funded by CAL FIRE (award number: 8GB18415) the US Forest Service (award number: 21-CS-11052021-201), and an incubation grant from the Data Science Strategic Research Initiative at California Polytechnic State University.  Thanks to Allan Hollander, Jim Thorne, Russ White, Ronny H\"{a}nsch, and the anonymous reviewers for their comments on the manuscript.



\section{Data Availability}
The code is available at \url{https://github.com/jonathanventura/urban-tree-detection} and the data is available at \url{https://github.com/jonathanventura/urban-tree-detection-data}.  Tree detection results are available at OSF:4S859 \citep{osf} and as an interactive map at \url{https://ufei.calpoly.edu/}.

\bibliographystyle{elsarticle-harv} 
\bibliography{urban-tree-detection}

\appendix

\section{HR-SFANet architecture}
\label{sec:pseudocode}

\new{Here we provide pseudo-code to exactly describe the architecture of our HR-SFANet network, which is a modification of the original SFANet architecture \cite{zhu2019dual}.  The pseudocode makes use of the following standard neural network operations:
\begin{itemize}
    \item $\textproc{Conv}(n,k,x)$: 2D convolution on input $x$ with a filter size of $k\times k$ and $n$ output channels.
    \item $\textproc{ReLU}(x) = \max(0,x)$: Element-wise rectified linear unit.
    \item $\textproc{Sigmoid}(x) = 1/(1+\exp(-x))$: Element-wise sigmoid function.
    \item $\textproc{BatchNorm}(x)$: Batch normalization \citep{ioffe2015batch}.
    \item $\textproc{MaxPool}(x)$: 2x2 max pooling with a stride of 2.
    \item $\textproc{Concatenate}(x,y)$: Channel-wise concatenation of $x$ and $y$.
    \item $\textproc{Upsample}(x)$: $2\times$ upsampling of input $x$ using bilinear interpolation.
\end{itemize}}
\begin{algorithm*}
\caption{High-Resolution SFANet}\label{alg:hr-sfanet}
\begin{algorithmic}
\Function{HR-SFANet}{input}
  \State conv12, conv22, conv33, conv43, conv53 $\gets \textproc{VGG16Backbone}(\textrm{input})$
  \State attention\_map $\gets \textproc{Decoder}$(conv12, conv22, conv33, conv43, conv53)
  \State attention\_map $\gets \textproc{Sigmoid}(\textproc{BatchNorm}(\textproc{Conv}(1,1,$attention\_map)))
  \State confidence\_map $\gets \textproc{Decoder}$(conv12, conv22, conv33, conv43, conv53)
  \State confidence\_map $\gets \textproc{Conv}(1,1,$ attention\_map $*$ confidence\_map)
\State \Return{confidence\_map, attention\_map}
\EndFunction
\end{algorithmic}
\end{algorithm*}

\begin{algorithm*}
\caption{VGG-16 Backbone}\label{alg:vgg16}
\begin{algorithmic}
\Function{VGG16Backbone}{input}
  \State $\textrm{conv11} \gets \textproc{ReLU}(\textproc{BatchNorm}(\textproc{Conv}(64,3,input))$
  \Comment{Block 1}
  \State $\textrm{conv12} \gets \textproc{ReLU}(\textproc{BatchNorm}(\textproc{Conv}(64,3,\textrm{conv11}))$
  
  \State $\textrm{conv21} \gets \textproc{ReLU}(\textproc{BatchNorm}(\textproc{Conv}(128,3,\textrm{\textproc{MaxPool}(conv21)}))$
  \Comment{Block 2}
  \State $\textrm{conv22} \gets \textproc{ReLU}(\textproc{BatchNorm}(\textproc{Conv}(128,3,\textrm{conv21}))$
  
  \State $\textrm{conv31} \gets \textproc{ReLU}(\textproc{BatchNorm}(\textproc{Conv}(256,3,\textproc{MaxPool}(\textrm{conv22})))$
  \Comment{Block 3}
  \State $\textrm{conv32} \gets \textproc{ReLU}(\textproc{BatchNorm}(\textproc{Conv}(256,3,\textrm{conv31}))$
  \State $\textrm{conv33} \gets \textproc{ReLU}(\textproc{BatchNorm}(\textproc{Conv}(256,3,\textrm{conv32}))$

  \State $\textrm{conv41} \gets \textproc{ReLU}(\textproc{BatchNorm}(\textproc{Conv}(512,3,\textrm{\textproc{MaxPool}(conv33)}))$
  \Comment{Block 4}
  \State $\textrm{conv42} \gets \textproc{ReLU}(\textproc{BatchNorm}(\textproc{Conv}(512,3,\textrm{conv41}))$
  \State $\textrm{conv43} \gets \textproc{ReLU}(\textproc{BatchNorm}(\textproc{Conv}(512,3,\textrm{conv42}))$

  \State $\textrm{conv51} \gets \textproc{ReLU}(\textproc{BatchNorm}(\textproc{Conv}(512,3,\textproc{MaxPool}(\textrm{conv43})))$
  \Comment{Block 5}
  \State $\textrm{conv52} \gets \textproc{ReLU}(\textproc{BatchNorm}(\textproc{Conv}(512,3,\textrm{conv51}))$
  \State $\textrm{conv53} \gets \textproc{ReLU}(\textproc{BatchNorm}(\textproc{Conv}(512,3,\textrm{conv52}))$
  
  \State \Return{conv12, conv22, conv33, conv43, conv53}
\EndFunction
\end{algorithmic}
\end{algorithm*}

\begin{algorithm*}
\caption{Decoder}\label{alg:decoder}
\begin{algorithmic}
\Function{Decoder}{conv11, conv22, conv33, conv43, conv53}
\State $x \gets \textproc{Concatenate}(\textproc{Upsample}(\textrm{conv53}),\textrm{conv43})$
\Comment{Block 1}
\State $x \gets \textproc{ReLU}(\textproc{BatchNorm}(\textproc{Conv}(256,1,x)))$
\State $x \gets \textproc{ReLU}(\textproc{BatchNorm}(\textproc{Conv}(256,3,x)))$

\State $x \gets \textproc{Concatenate}(\textproc{Upsample}(x),\textrm{conv33})$
\Comment{Block 2}
\State $x \gets \textproc{ReLU}(\textproc{BatchNorm}(\textproc{Conv}(128,1,x)))$
\State $x \gets \textproc{ReLU}(\textproc{BatchNorm}(\textproc{Conv}(128,3,x)))$

\State $x \gets \textproc{Concatenate}(\textproc{Upsample}(x),\textrm{conv22})$
\Comment{Block 3}
\State $x \gets \textproc{ReLU}(\textproc{BatchNorm}(\textproc{Conv}(64,1,x)))$
\State $x \gets \textproc{ReLU}(\textproc{BatchNorm}(\textproc{Conv}(64,3,x)))$
\State $x \gets \textproc{ReLU}(\textproc{BatchNorm}(\textproc{Conv}(32,3,x)))$

\State $x \gets \textproc{Concatenate}(\textproc{Upsample}(x),\textrm{conv12})$
\Comment{Block 4}
\State $x \gets \textproc{ReLU}(\textproc{BatchNorm}(\textproc{Conv}(32,1,x)))$
\State $x \gets \textproc{ReLU}(\textproc{BatchNorm}(\textproc{Conv}(32,3,x)))$
\State $x \gets \textproc{ReLU}(\textproc{BatchNorm}(\textproc{Conv}(32,3,x)))$
\State \Return{$x$}
\EndFunction
\end{algorithmic}
\end{algorithm*}

\end{document}